\documentclass{article}

\usepackage[preprint]{neurips_2021}

\usepackage[utf8]{inputenc} 
\usepackage[T1]{fontenc}    
\usepackage{hyperref}       
\usepackage{url}            
\usepackage{booktabs}       
\usepackage{amsfonts}       
\usepackage{nicefrac}       
\usepackage{microtype}      
\usepackage{xcolor}         
\usepackage{float}
\usepackage{graphicx}
\usepackage{listings}
\usepackage{multirow}
\usepackage{subcaption}
\usepackage{enumitem}
\usepackage{amssymb,amsmath,amsthm}
\usepackage[nounderscore]{syntax}
\usepackage{wrapfig}

\newtheorem{definition}{Definition}

\definecolor{pltblue}{HTML}{1f77b4}
\definecolor{pltorange}{HTML}{ff7f0e}

\newcommand\blfootnote[1]{%
  \begingroup
  \renewcommand\thefootnote{}\footnote{#1}%
  \addtocounter{footnote}{-1}%
  \endgroup
}

\renewcommand\vec{\mathbf}

\newcommand{\system}{MISIM}

\title{MISIM: A Neural Code Semantics Similarity System \\ Using the Context-Aware Semantics Structure}

\author{%
  Fangke Ye $^*$\\
  Intel Labs and Georgia Institute of Technology\\
  \footnotesize{\texttt{yefangke@gatech.edu}}
  \And
  Shengtian Zhou $^*$\\
  Intel Labs\\
  \footnotesize{\texttt{shengtian.zhou@intel.com}}
  \And
  Anand Venkat\\
  Intel Labs\\
  \footnotesize{\texttt{anand.venkat@intel.com}}
  \And
  Ryan Marcus\\
  Intel Labs and MIT\\
  \footnotesize{\texttt{ryanmarcus@csail.mit.edu}}
  \And
  Nesime Tatbul\\
  Intel Labs and MIT \\
  \footnotesize{\texttt{tatbul@csail.mit.edu}}
  \And
  Jesmin Jahan Tithi\\
  Intel Labs\\
  \footnotesize{\texttt{jesmin.jahan.tithi@intel.com}}
  \And
  Niranjan Hasabnis\\
  Intel Labs\\
  \footnotesize{\texttt{niranjan.hasabnis@intel.com}}
  \And
  Paul Petersen\\
  Intel\\
  \footnotesize{\texttt{paul.petersen@intel.com}}
  \And
  Timothy Mattson\\
  Intel Labs\\ 
  \footnotesize{\texttt{timothy.g.mattson@intel.com}}
  \And
  Tim Kraska\\
  MIT\\
  \footnotesize{\texttt{kraska@mit.edu}}
  \And
  Pradeep Dubey\\
  Intel Labs\\
  \footnotesize{\texttt{pradeep.dubey@intel.com}}
  \And
  Vivek Sarkar\\
  Georgia Institute of Technology\\
  \footnotesize{\texttt{vsarkar@gatech.edu}}
  \And
  Justin Gottschlich\\
  Intel Labs and University of Pennsylvania\\
  \footnotesize{\texttt{justin.gottschlich@intel.com}}
}

\begin{document}

\maketitle

\blfootnote{$^*$ Lead authors.}

\begin{abstract}

Code semantics similarity can be used for many tasks such as code recommendation, automated software defect correction, and clone detection. Yet, the accuracy of such systems has not yet reached a level of general purpose reliability. To help address this, we present \emph{\underline{M}achine \underline{I}nferred Code \underline{Sim}ilarity} (MISIM), a neural code semantics similarity system consisting of two core components: \emph{(i)} MISIM uses a novel \emph{context-aware semantics structure}, which was purpose-built to lift semantics from code syntax; \emph{(ii)} MISIM uses an extensible neural code similarity scoring algorithm, which can be used for various neural network architectures with learned parameters. We compare MISIM to four state-of-the-art systems, including two additional hand-customized models, over 328K programs consisting of over 18 million lines of code. Our experiments show that MISIM has $8.08\%$ better accuracy (using MAP@R) compared to the next best performing system.

\end{abstract}

\section{Introduction} \label{sect:intro}

\vspace{-5pt}

The field of \emph{machine programming} (MP) is concerned with the automation of software development~\citep{gottschlich:2018:mapl}. In recent years, there has been an emergence of many MP systems due, in part, to advances in machine learning, formal methods, data availability, and computing efficiency~\citep{allamanis:2018:acm, alon:2018:pldi, alon:2019:popl, alon:2019:iclr, ben-nun:2018:neurips, cosentino:2017:ieee, li:2017:icsme, luan:2019:oopsla, odena:2020:iclr, program_synthesis:2020:anl, tufano:2018:msr, wei:2017:ijcai, zhang:2019:icse, zhao:2018:esec/fse}. An open challenge in MP is in the construction of accurate code similarity systems. \emph{Code similarity}, which determines if two or more code fragments are similar, can be reasoned about in many ways. Two principle ways are through syntactic similarity and semantic analysis. While attention has historically centered around \emph{code syntax similarity} (i.e., whether two or more code fragments are syntactically similar) recent work has revealed many advantages of \emph{code semantics similarity} (i.e., whether two or more code fragments are similar in meaning, even in the presence of syntactic differences)~\citep{iyer:2020:cap, lee:2021:code}. Precisely, we define two code fragments $C_i$ and $C_j$ to be \emph{semantically equivalent} if for a given set of inputs, $I$, both $C_i$ and $C_j$, produce an identical respective set of outputs, $O$. 

\begin{figure*}[ht]
\begin{center}
\includegraphics[width=\textwidth]{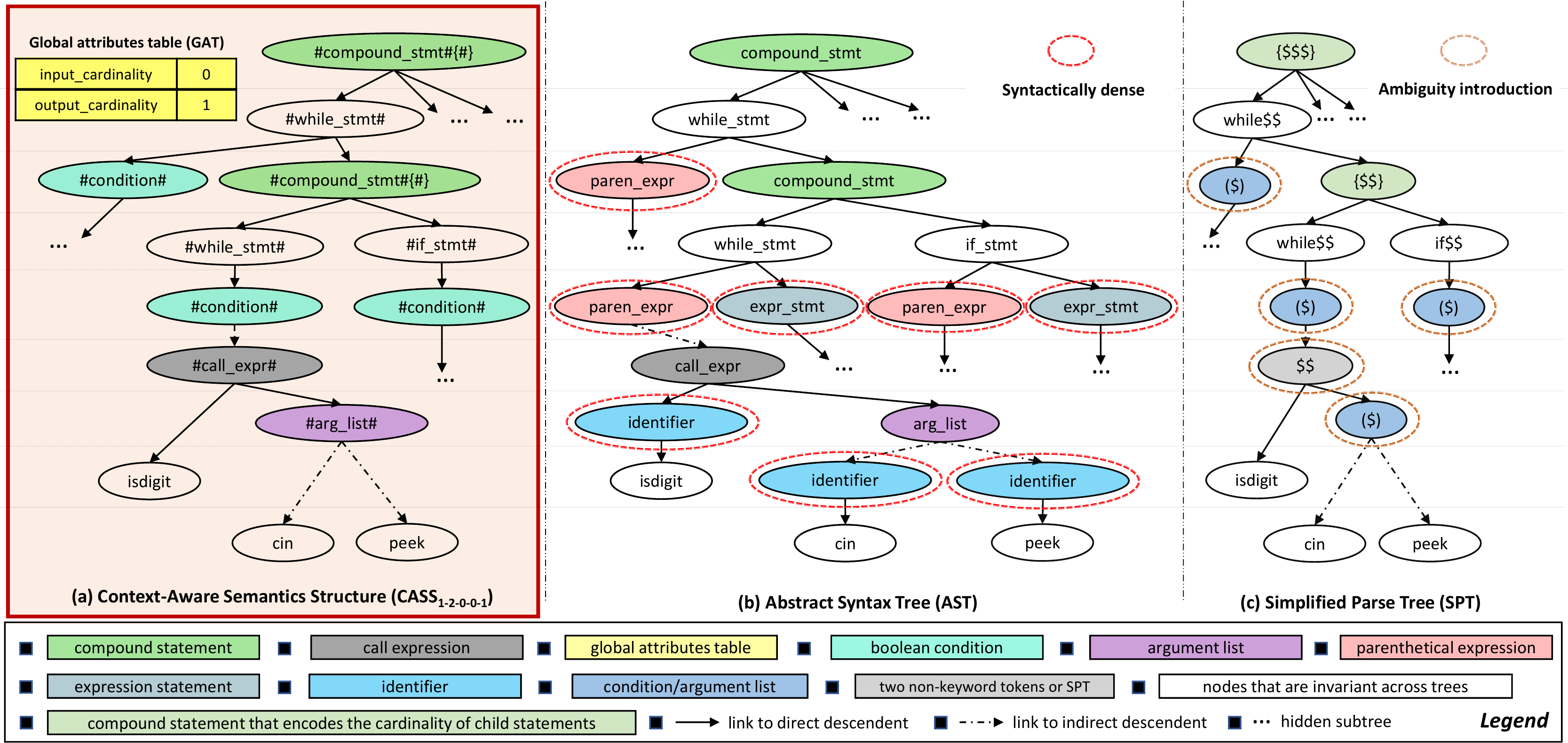}
\vspace{-5pt}
\caption{Three structural representations of Program A: CASS, AST, and SPT.}
\label{fig:motivation}
\end{center}
\vspace{-10pt}
\end{figure*}

Code semantics similarity systems aim to determine if code fragments are solving a similar (or equivalent) problem, even if implemented differently (e.g., various algorithms of {\footnotesize{\texttt{sort()}}}~\citep{cormen:2009:algo}). They can be used in many ways. One way is to improve programmer productivity on tasks such as code recommendation, clone detection, bug detection and mitigation, and language-to-language translation (i.e., transpilation), to name a few~\citep{allamanis:2018:iclr, ahmad:2019:siggraph, bader:2019:oopsla, barman:2016:oopsla, bhatia:2018:icse, dinella:2020:iclr, kamil:2016:pldi, luan:2019:oopsla, pradel:2018:oopsla}. Moreover, with improved accuracy, code semantics similarity systems can likely be leveraged to automate many other parts of software development, which may become necessary to keep pace with the heterogeneous growth of programming languages and hardware systems~\citep{ahmad:2019:siggraph, batra:2018:mckinsey, bogdan:2019:codes, chen:2020:sd, deng:2020:ieee, hannigan:2019:mckinsey}. Yet, despite advances in the space of code semantics similarity, fundamental questions remain open. One open question, which is the principle focus of this paper, is regarding {\em how the structural representations impact learning semantics similarity in syntactically diverse code sets}~\citep{alam:2019:neurips, allamanis:2018:iclr, becker:2017:corr, ben-nun:2018:neurips, dinella:2020:iclr, iyer:2020:cap, luan:2019:oopsla}.

While prior work has explored some structural representations of code in the space of code similarity and understanding, these explorations are still in their early stages. The classical abstract syntax tree (AST) is used in the code2vec and code2seq systems~\citep{alon:2019:popl, alon:2019:iclr}, while two novel structures called the conteXtual flow graph (XFG) and the simplified parse tree (SPT) are used in Neural Code Comprehension (NCC)~\citep{ben-nun:2018:neurips} and Aroma~\citep{luan:2019:oopsla}, respectively. While each of these representations has benefits in certain contexts, they possess one or more limitations when considered more broadly. For example, the AST -- while having a notable historical importance for optimizing compilers -- can be syntactically dense (see Figure ~\ref{fig:motivation}(b)). Such syntax density can mislead code similarity systems into memorizing syntax, rather than learning semantics (i.e., the meaning behind the syntax). Alternatively, the XFG can capture important data dependencies, but is obtained from an intermediate representation (IR) that requires code compilation. This restriction can limit its application in interactive developer environments such as live code auto-completion or recommendation. The SPT is structurally driven, which enables it to lift certain semantic meaning from code, yet, it does not always resolve syntactic ambiguities. Instead, it can sometimes introduce them, due to its intentional coarse-grain approach (see Figure~\ref{fig:motivation}(c)).

Learning from these observations, we present our code semantics similarity system called \emph{\underline{M}achine \underline{I}nferred Code \underline{Sim}ilarity} (MISIM). We principally focus on MISIM's two main novelties: \emph{(i)} its novel code structural representation, the \emph{context-aware semantics structure} (CASS), and \emph{(ii)} its neural-based \emph{learned} code similarity scoring algorithm. This paper makes the following technical contributions:

\begin{itemize}
[nosep,leftmargin=1em,labelwidth=*,align=left]

\item We present \system's \emph{context-aware semantics structure} (CASS), a
configurable representation of code designed to \emph{(i)} lift semantic
meaning from code syntax, which \emph{(ii)} supports language-specific and language-agnostic extensibility.

\item We present \system's learned deep neural network (DNN) semantics similarity scoring framework and show its efficacy across three DNN topologies: \emph{(i)} bag-of-features, \emph{(ii)} a recurrent neural network (RNN), and \emph{(iii)} a graph neural network (GNN).

\item We compare MISIM to four existing code similarity systems: \emph{(i)} code2vec, \emph{(ii)} code2seq, \emph{(iii)} Neural Code Comprehension, and \emph{(iv)} Aroma. To deepen the experimental analysis, we also include two customized source code token sequence models and include reported results from IBM and MIT's CodeNet repository, who conducted their own comparison of MISIM to Aroma. Across approximately 18 million lines of code (not including CodeNet), our results show that \system\ has, at its worst, $8.08\%$ better accuracy than the next best performing system.

\end{itemize}


\section{Code Representations for Code Semantics Similarity Analysis}
\label{sec:motivating_example}

Code representation is a core component of code semantics similarity analysis. Existing approaches fall under two categories: syntax-based representations and semantics-based representations. We provide a brief anecdotal analysis of these representations, discuss their strengths and weaknesses for code semantics similarity, and motivate a new approach that offers more configurability.

{\bf A Real-World Example.}
We analyze two simple code examples (Program A and B) taken from the POJ-104 dataset~\citep{mou:2016:aaai}. These programs correctly solve the same problem (\#88), where the goal is to emit all digits in a given input string. Hence, while the implementations of Program A and B are syntactically dissimilar, they are semantically equivalent.
Code for Program A and B is shown below;
Figure~\ref{fig:motivation} illustrates three different ways to represent Program A.

\lstdefinestyle{codestyle}{frame=single, columns=fullflexible, basicstyle={\scriptsize\ttfamily}}
\lstset{style=codestyle}
\noindent\begin{minipage}{.45\textwidth}
\begin{lstlisting}[language=C++]
Program A

  int a;
  // algorithm
  while (!cin.eof()) {
    while (!cin.eof() && !isdigit(cin.peek()))
      cin.get(); // ignore
    // print out result
    if (cin >> a) 
        cout << a << endl;
  }
\end{lstlisting}
\end{minipage}\hfill
\noindent\begin{minipage}{.51\textwidth}
\begin{lstlisting}[language=C++]
Program B

  char *p, *head, c;
  p = (char *) malloc(sizeof(char) * 30);
  head = p; scanf("%c", p);
  while (*p != '\n') { p++; *p = getchar();}
  *p = '\0'; p = head;
  for (; *p != '\0'; p++) {
    if(*p <= '9' && *p >= '0'){printf("%c",*p);}
    else if(*(p+1) < 58 && *(p+1) > 47){putchar('\n');}
  } 
\end{lstlisting}
\end{minipage}

{\bf Syntax-based Representations.}
Compilers have successfully used syntax-based representations of programs, such as parse trees (a.k.a., concrete syntax trees) and abstract syntax trees (ASTs), for several decades~\citep{baxter:1998:icsme}. More recently, the code understanding systems {\footnotesize{\texttt{code2vec}}} and {\footnotesize{\texttt{code2seq}}} have utilized ASTs as their basic representation of code for program semantics analysis~\citep{alon:2019:popl, alon:2019:iclr}.
Parse trees are typically built by a language-specific parser during the compilation process and faithfully capture every syntactic detail of the source program.
ASTs, on the other hand, abstract away certain syntactic details (e.g., parentheses), which can simplify program analysis. However, ASTs can still be syntactically dense, which can mislead code similarity systems into memorizing syntax rather than learning semantics. For instance, the AST in Figure~\ref{fig:motivation}(b) represents the parentheses of {\footnotesize{\texttt{while}}} statements in Program A as {\footnotesize{\texttt{paren\_expr}}} nodes, while missing the semantic binding they have to the condition expression of the {\footnotesize{\texttt{while}}} statements. Syntax-based representations are also conceptually ill-fit for cross-language code semantics similarity in general.

{\bf Semantics-based Representations.}
Semantics-based representations capture the semantics of various program constructs instead of syntax. ConteXtual flow graph (XFG) used by NCC and simplified parse tree (SPT) used by Aroma are two such representations. NCC hypothesizes that program statements that operate in similar contexts are semantically similar, where the context of a statement is defined as the other surrounding statements with direct data- and control-flow dependencies. These dependencies are captured at the IR level by defining programs' XFG representations. {\footnotesize{\texttt{inst2vec}}} embeddings are then trained for IR instructions a la {\footnotesize{\texttt{word2vec}}}~\citep{mikolov:2013:iclr} and skip-gram models~\citep{mikolov:2013:nips}. Overall, NCC views instructions as semantically similar, if their embedding vectors are closer. Unfortunately, NCC's dependency on a compiler IR restricts XFG only to compilable code. Aroma uses SPT as a program representation to enable ML-based code search and recommendation.
Aroma builds SPT of a given query program, featurizes it using a set of manually-selected features, and performs dot product of the query's feature vector and that of the candidates to find similar code snippets. SPT is different than AST in that, it only consists of program tokens and does not use any special language-specific rule names. Thus, SPT is language-agnostic, which enables uniform handling of code written in different programming languages.

{\bf Limitations.}
While semantics-based code representations, and SPT in particular, are more suitable for general-purpose code semantics similarity analysis, our example reveals some critical limitations with SPT. 
First of all, although SPT is structurally-driven (as opposed to the syntax-driven AST~\citep{luan:2019:oopsla}), it may unintentionally carry {\em semantic ambiguity}. For instance, the SPT in Figure~\ref{fig:motivation}(c) does not distinguish between the argument list of a function call (e.g., "{\footnotesize{\texttt{(cin.peek())}}}") and the condition node of an {\footnotesize{\texttt{if}}} statement (e.g., "{\footnotesize{\texttt{(cin >{}> a)}}}"), and represents both of them by "{\footnotesize{\texttt{(\$)}}}". Moreover, it also captures program details that may be irrelevant for code semantics similarity. For instance, it captures the number of program statements in {\footnotesize{\texttt{main}}} in the root node (e.g., as 3 {\footnotesize{\texttt{\$'s}}} for Program A, and 8 {\footnotesize{\texttt{\$'s}}} for Program B), yet, while both of these programs have different number of statements, they are semantically equivalent. Therefore, featurizing on this metric could mislead and mistrain an ML system to infer such information as semantically meaningful.

{\bf A New Approach.}
We propose a new way to represent code, called the \emph{context-aware semantics structure} (CASS), to improve upon the limitations of the existing representations. CASS is a semantics-based representation that builds on and extends SPT into a configurable structure, which enables it to flexibly capture a wide variety of structural representations (details in Section~\ref{sect:system}). Figure~\ref{fig:motivation}(a) shows the CASS for Program A, obtained from one of its 216 different configurations. In this example, CASS resolves the aforementioned ambiguity introduced by SPT by representing argument list and condition node of an {\footnotesize{\texttt{if}}} and {\footnotesize{\texttt{while}}} statement differently, while simultaneously eliding away unnecessary syntactic density produced by the AST (e.g., {\footnotesize{\texttt{identifier}}} nodes). These modifications help in building a more accurate neural backend system (e.g., for this particular case, CASS-based MISIM outperforms both its AST- and SPT-based counterparts
by more than a 4.3\% margin). 




\section{The \system \ System} \label{sect:system}

\begin{figure*}[ht]
\vspace{-5pt}
\begin{center}
\includegraphics[width=\textwidth]{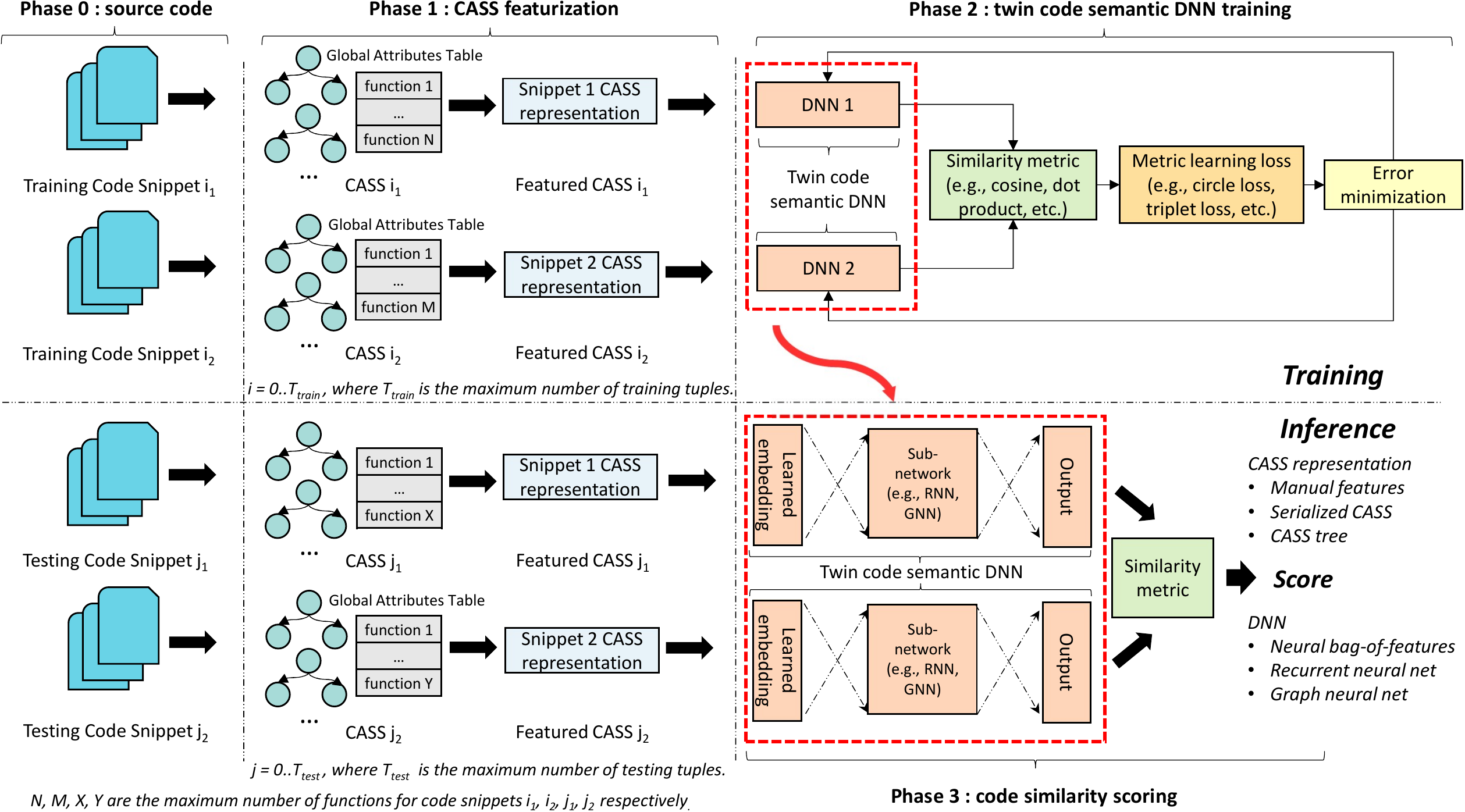}
\vspace{-5pt}
\caption{Overview of the \system{} system.}
\label{fig:system_overview}
\end{center}
\vspace{-10pt}
\end{figure*}

In Figure~\ref{fig:system_overview}, we provide an overview of the \system\ system. A core component of \system\ is the novel \emph{context-aware semantics structure} (CASS), which aims to capture semantically salient properties of the input code. Moreover, CASS is designed to be \textit{context-aware}. That is, it can capture information that describes the context of the code (e.g., parenthetical operator disambiguation between a function call, mathematical operator precedence, Boolean logic ordering, etc.) that may otherwise be ambiguous without such context-sensitivity. Once these CASSes are constructed, they are vectorized and used as input to a neural network, which produces a feature vector for a corresponding CASS. Once a feature vector is generated, a code similarity measurement (e.g., cosine similarity~\citep{baeza-yates:1999:longman}) calculates the similarity score.

\subsection{Context-Aware Semantics Structure (CASS)}
\label{sec:cast}

We have designed CASS with the following guiding principles: \emph{\textbf{(i)}} it should not require compilation, \emph{\textbf{(ii)}} it should be a flexible representation that captures code semantics, and \emph{\textbf{(iii)}} it should be capable of resolving code ambiguities in both its context sensitivity to the code and its environment. The first principle \emph{\textbf{(i)}} originates from the observation that unlike programs written in higher-level scripting languages (e.g., Python~\citep{rossum:2009:python3}, JavaScript~\citep{flanagan:2006:javascript}), C/C++ programs found ``in the wild'' may not be well-formed (e.g., due to specialized compiler dependencies) or exhaustively include all of their dependencies (e.g., due to assumptions about library availability) and therefore may not compile. Moreover, for code recommendation systems that are expected to function in a live setting, requiring compilation may severely constrain their practical application. We address this by introducing a structure such that it does not require compilation (Section~\ref{sect:treeandgat}). The second \emph{\textbf{(ii)}} and third \emph{\textbf{(iii)}} principles originate from the observation that different scenarios may require attention to different semantics (e.g., embedded memory-bound systems may prefer to use algorithms that do not use recursion due to a potential call stack overflow) and that programming languages (PLs) evolve and new PLs continue to emerge. We attempt to address these issues with CASS's configuration categories (Section~\ref{sect:categories}).

\begin{wraptable}{r}{180pt}
\setlength\tabcolsep{2pt}
\vspace{-10pt}
\caption{CASS Configuration Options.}
\vspace{-5pt}
\centering
\scriptsize
\begin{tabular}{|c|l|}
\hline
\multicolumn{2}{|c|}{\textbf{Language-specific}} \\ \hline
Type                                 & Option                                            \\\hline
\multirow{4}{1.5cm}{Node Prefix Label}    & 0. No change (original SPT)  \\
                                         & 1. Add a prefix to each internal node label    \\
                                         & 2. Add a prefix to parenthesis node label \\
                                         & \hspace{0.25cm}(C/C++ Specific) \\
\hline \hline
\multicolumn{2}{|c|}{\textbf{Language-agnostic}} \\ \hline
Type                                 & Option                                            \\\hline
\multirow{4}{1.5cm}{Compound Statements} & 0. No change (original SPT) \\
                                        & 1. Drop all features relevant to compound \\
                                        & \hspace{0.25cm} statements  \\
                                        & 2. Replace with `\{\#\}'                                \\\hline
\multirow{3}{1.5cm}{Global Variables} & 0. No change (original SPT) \\
                                     & 1. Drop all features relevant to global vars     \\
                                     & 2. Replace with `\#GVAR'                                \\
                                     & 3. Replace with `\#VAR' (label for local vars) \\\hline
\multirow{4}{1.5cm}{Global Functions} & 0. No change (original SPT)            \\
                                     & 1. Drop all features relevant to global functions     \\
                                     & 2. Drop function identifier and replace with\\
                                     & \hspace{0.25cm} `\#EXFUNC' \\\hline
\multirow{3}{1.5cm}{Function I/O Cardinality} & 0. No change                             \\
                                     & 1. Include the input and output cardinality \\
                                     & \hspace{0.25cm} per function in GAT \\\hline
\end{tabular}%
\label{tab:configurations}
\vspace{-15pt}
\end{wraptable}

\subsubsection{CASS tree and global attributes table}
\label{sect:treeandgat}
Here we provide an informal definition of CASS (a formal definition is in Appendix~\ref{appendix:cass}). The CASS consists of one or more CASS trees and an optional global attributes table (GAT). A CASS tree is a tree, in which the root node represents the entire span of the code snippet. During the construction of a CASS tree, the program tokens are mapped to their corresponding node labels using the grammar of the high-level programming language. A CASS's GAT contains exactly one entry per unique function definition in the code snippet. A GAT entry currently includes only the input and output cardinality values for each corresponding function, but can be extended as new global attributes are needed. 



\subsubsection{CASS configuration categories}

\label{sect:categories}
In general, CASS configurations can be broadly classified into two categories: \emph{language-specific} and \emph{language-agnostic}. Exact values of the options for each of the configuration categories are described in Table~\ref{tab:configurations}. Below we provide an intuitive description of the categories and their values.

\textbf{Language-specific configurations (LSCs).} Language-specific configurations are meant to capture semantic meaning by resolving syntactic ambiguities that could be present in the concrete syntax trees. It also introduces specificity related to the high-level programming language. For example, the parentheses operator is overloaded in many programming languages to enforce an order of evaluation of operands as well as to enclose a list of function arguments, amongst other things. CASS disambiguates these by explicitly embedding the semantic contextual information in the CASS tree nodes using the \emph{node prefix label} (defined in Appendix~\ref{appendix:cass}).

\textbf{\textit{(A) Node Prefix Label.}} The configuration options for node prefix labels\footnote{Analytically deriving the optimal selection of node prefix labels across all C/C++ code may be untenable. To accommodate this, we currently provide two levels of granularity for C/C++ node prefix labels in CASS.} correspond to various levels of semantic to syntactic information. In Table~\ref{tab:configurations}, option 0 corresponds to the extreme case of a concrete syntax embedding, option 1 corresponds to eliminating irrelevant syntax, and option 2 is principally equivalent to option 1, except it applies only to parentheticals, which we have identified -- through empirical evaluation -- to often have notably divergent semantic meaning based on context.

\textbf{Language-agnostic configurations (LACs).} LACs can improve code similarity analysis by unbinding overly-specific semantics that may be present in the original concrete syntax tree structure.

\textbf{\textit{(B) Compound Statements.}}
The \textit{compound statements} configuration option enables the user to control how much non-terminal node information is incorporated into the CASS. Option 0 is equivalent to Aroma's SPT, option 1 omits separate features for compound statements altogether, and option 2 does not discriminate between compound statements of different lengths and specifies a special label to denote the presence of a compound statement.

\textbf{\textit{(C) Global Variables.}}
The \textit{global variables} configuration specifies the degree of global variable-specific information contained in a CASS. In other words, it provides the user with the ability to control the level of abstraction -- essentially binding or unbinding global variable names as needed. If all code similarity analysis will be performed against the same software program, retaining global variable names may help elicit deeper semantic meaning. If not, unbinding global variable names may improve semantic meaning.

\textbf{\textit{(D) Global Functions.}}
The \textit{global functions} configuration serves the dual purpose of \emph{(i)} controlling the amount of function-specific information to featurize and \emph{(ii)} to explicitly disambiguate between the usage of global functions and global variables (a feature that is absent in Aroma's SPT design).

\textbf{\textit{(E) Function I/O Cardinality.}}
The \textit{function I/O cardinality} configuration aims to abstract the semantics of certain groups of functions through input and output cardinality (i.e., embedded semantic information that can be implicitly derived by analyzing the number of input and output parameters of a function). I/O cardinality values of every function are recorded in GAT, if this option is enabled.

We have found that the specific context in which code similarity is performed provides an indication of the optimal CASS configuration. We discuss this in greater detail in Appendix~\ref{subsec:intuit}.

\subsection{Neural Scoring Algorithm}

\system's neural scoring algorithm aims to compute the similarity score of two input programs. The algorithm consists of two phases. The first phase involves a neural network model that maps a featurized CASS to a real-valued code vector. The second phase generates a similarity score between a pair of code vectors using a similarity metric.\footnote{For this work, we have chosen cosine similarity as the similarity metric used within \system.} We describe the details of the scoring model, its training strategy, and other neural network model choices in this section.

\subsubsection{Model}

We investigated three neural network approaches for \system's scoring algorithm:
\emph{(i)} a graph neural network (GNN), \emph{(ii)} a
recurrent neural network (RNN), and \emph{(iii)} a bag of manual features (BoF)
neural network. We name these models \system-GNN, \system-RNN, and \system-BoF
respectively.
The graphical nature of CASS, as well as the recent success in applying GNNs in the program domain~\cite{allamanis:2018:iclr,brockschmidt:2019:iclr,dinella:2020:iclr,wei:2020:iclr}, leads us to design the \system-GNN model that directly encodes the graphical structure of CASS.
\system-RNN is based on~\cite{hu:2018:icpc}, which serializes a CASS into a sequence and uses an RNN to encode the structure.%
\footnote{We did not include a Transformer-based model using the serialized CASS because in our experiments, we had observed consistently worse performance from Transformers than that from RNNs when applying them to source code token sequences.}
Unlike the two aforementioned models, \system-BoF takes in not the CASS but a bag of manual features extracted from it, and uses a feed-forward network to encode them into a vector.
We compared the three models in our experiments and observed that \system-GNN performed the best overall.
Therefore, we describe it in detail in this section. Appendix~\ref{appendix:models} has details of the \system-RNN and \system-BoF models.

\textbf{\system-GNN.} In the \system-GNN model, an input program's CASS representation
is transformed into a graph. Then, each node in the graph is embedded into a
trainable vector, serving as the node's initial state. Next, a GNN is used to
update each node's state iteratively. Finally, a global readout function is
applied to extract a vector representation of the entire graph from the 
final states of the nodes. We describe each of these steps in more detail below.


\textbf{Input Graph Construction.}
We represent each program as a single CASS instance. Each instance can contain
one or more CASS trees, where each tree corresponds to a unique function of the
program. The CASS instance is converted into a single graph representation to
serve as the input to the model. The graph is constructed by first transforming
each CASS tree and its GAT entry into an individual graph. These graphs are then
merged into a single (disjoint) graph.
For a CASS consisting of a CASS tree $T=(V,E)$ and a GAT entry $a$, we transform it
into a directed graph $G=(V',E',R)$, where $V'$ is the set of graph nodes, $R$
is the set of edge types, and $E'=\{ (v, u, r) \mid v, u \in V', r \in R \}$ is
the set of graph edges. The graph is constructed as follows:
\vspace{-2pt}
\begin{gather*}
V' = V \cup \{ a \}, \quad
R = \{ p, c \}, \quad
E' = \{ (v, u, p) \mid (v, u) \in E \} \cup \{ (v, u, c) \mid (u, v) \in E \}.
\end{gather*}
\vspace{-2pt}
The two edge types, $p$ and $c$, represent edges from CASS tree nodes to their
parent and children nodes, respectively.

\textbf{Graph Neural Network.}
\system \  embeds each node $v \in V'$ in the input graph $G$ into a vector by
assigning a trainable vector to each unique node label (with the optional
prefix) and GAT attribute. The node embeddings are then used as node initial
states ($\vec{h}_v^{(0)}$) by a relational graph convolutional network
(R-GCN~\citep{schlichtkrull:2018:eswc}) specified as the following:
\begin{footnotesize}
\begin{align*}
\vec{h}_v^{(l)} =
\mathrm{ReLU} \Bigg( &
    \frac{1}{ \sum_{r \in R} |\mathcal{N}_v^r| } \sum_{r \in R} \sum_{u \in \mathcal{N}_v^r}
        \vec{W}_r^{(l)} \vec{h}_u^{(l-1)}
    +
    \vec{W}_0^{(l)} \vec{h}_v^{(l-1)}
\Bigg)
\quad\quad
v \in V', l \in [1, L]
,
\end{align*}
\end{footnotesize}
where $L$ is the number of GNN layers, $\mathcal{N}_v^r = \{ u \mid (u, v, r)
\in E' \}$ is the set of neighbors of $v$ that connect to $v$ through an edge of
type $r \in R$, and $\vec{W}_r^{(l)}$, $\vec{W}_0^{(l)}$ are weight matrices to
be learned.

\textbf{Code Vector Generation.}
To obtain a code vector $\vec{c}$ that represents the entire input graph,
we apply a graph-level readout function as specified below:
\begin{footnotesize}
\begin{align*}
\vec{c} = \mathrm{FC} \left( \left[
\mathrm{AvgPool} \left(\left\{ \vec{h}_v^{(L)} \;\middle|\; v \in V' \right\}\right) ;
\mathrm{MaxPool} \left(\left\{ \vec{h}_v^{(L)} \;\middle|\; v \in V' \right\}\right)
\right] \right)
\end{align*}
\end{footnotesize}
The output vectors of average pooling and max pooling on the nodes' final states are concatenated and fed into a fully-connected layer, yielding the code vector for the entire input program.

\subsubsection{Training}
\label{sec:misim_training}

We train the neural network model following the setting of metric learning~\citep{schroff:2015:cvpr,hermans:2017:dblp,musgrave:2020:arxiv,sun:2020:cvpr}, which tries to map input data to a vector space where, under a distance (or similarity) metric, similar data points are close together (or have large similarity scores) and dissimilar data points are far apart. The metric we use is the cosine similarity in the code vector space. As shown in the lower half of Figure~\ref{fig:system_overview}, we use pair-wise labels to train the model. Each pair of input programs are mapped to two code vectors by the model, from which a similarity score is computed and optimized using a metric learning loss function.

\section{Experimental Evaluation} \label{sect:experiments}

In this section, we analyze the performance of \system{} compared to
code2vec, code2seq, NCC, and Aroma on two datasets containing a total of more than 328,000 programs.%
\footnote{
Although other code similarity systems exist, we were not able to compare to them due to the differences in target languages, problem settings, and lack of open-source availability.
}
We also compare it against our own hand-tuned and best performing recurrent neural network (Seq-RNN) and Transformer (Seq-Transformer), which take tokenized source code as input (see Appendix \ref{appendix:seq_models} for details).
Overall, we find that \system{} has greater accuracy than these systems across two metrics. We also perform an abbreviated analysis of two \system{} variants, each trained with a different CASS configuration, to provide insight into when different configurations may be better fit for different code corpora.

\textbf{Datasets.}
Our experiments are conducted on two datasets: the Google Code Jam (GCJ)
dataset~\citep{ullah:2019:access} and the POJ-104
dataset~\citep{mou:2016:aaai}. The GCJ dataset consists of solutions to programming problems in Google's Code Jam coding competitions.
We use a subset of it that consisting of C/C++ programs that solve 297 problems. The POJ-104 dataset consists of student-written C/C++ programs solving 104 problems. For both datasets, we label two programs as similar if they are solutions to the same problem. After a filtering step, which removes unparsable/non-compilable programs, we split each dataset by problem into
three subsets for \emph{training}, \emph{validation}, and \emph{testing}. Detailed statistics of the dataset partitioning are shown in Table~\ref{tab:dataset_stats}.
\begin{wraptable}{r}{170pt}
\setlength\tabcolsep{2pt}
\caption{Dataset statistics.}
\vspace{-5pt}
\label{tab:dataset_stats}
\centering
\scriptsize
\begin{tabular}{lcccc}
\toprule
\multirow{2}{*}{Split} & \multicolumn{2}{c}{GCJ} & \multicolumn{2}{c}{POJ-104} \\ 
\cmidrule(l{1.5pt}r{1.5pt}){2-3} \cmidrule(l{1.5pt}){4-5}
                       & \#Problems & \#Programs & \#Problems   & \#Programs   \\
\cmidrule(r{1.5pt}){1-1} \cmidrule(l{1.5pt}r{1.5pt}){2-3} \cmidrule(l{1.5pt}){4-5}
Training               & 237        & 223,171    & 64           & 28,137       \\
Validation             & 29         & 36,409     & 16           & 7,193        \\
Test                   & 31         & 22,795     & 24           & 10,450       \\ 
\cmidrule(r{1.5pt}){1-1} \cmidrule(l{1.5pt}r{1.5pt}){2-3} \cmidrule(l{1.5pt}){4-5}
Total                  & 297        & 282,375    & 104          & 45,780       \\
\bottomrule
\end{tabular}%
\vspace{-10pt}
\end{wraptable}

\textbf{Training.}
Unless otherwise specified, we use the same training procedure in all experiments. The models are built and trained using PyTorch~\citep{paszke:2019:neurips}. To train the models, we use the Circle loss~\citep{sun:2020:cvpr}, a state-of-the-art metric learning loss function that has been tested effective in various similarity learning tasks. Following the P-K sampling
strategy~\citep{hermans:2017:dblp}, we construct a batch of programs by first randomly sampling 16 different problems, and then randomly sampling at most 5 different solutions for each
problem. The loss function takes the similarity scores of all intra-batch pairs
and their pair-wise labels as input. Further details about the training procedure are in Appendix~\ref{appendix:training}.

%

\textbf{Evaluation Metrics.}
The accuracy metrics we use for evaluation are Mean Average Precision at R
(MAP@R)~\citep{musgrave:2020:arxiv} and Average Precision
(AP)~\citep{baeza-yates:1999:longman}. Since these metrics are already
defined, we do not detail them here (Details in Appendix~\ref{appendix:evaluation_metrics}.)

\begin{figure*}[t]
\vspace{-5pt}
\centering
\begin{subfigure}[t]{0.24\linewidth}
\includegraphics[width=\linewidth]{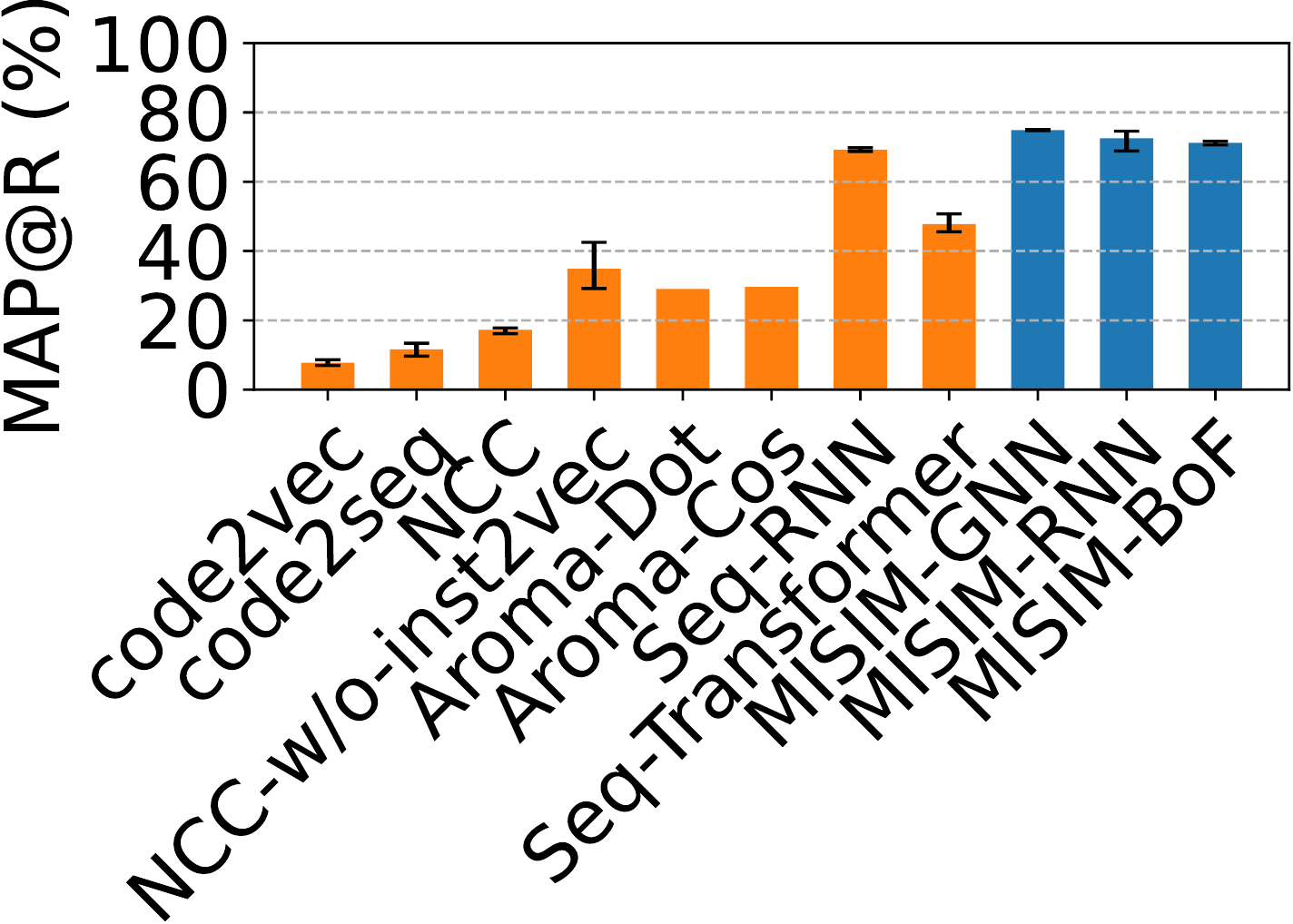}
\vspace{-15pt}
\caption{MAP@R on GCJ.}
\end{subfigure}
\hfill
\begin{subfigure}[t]{0.24\linewidth}
\includegraphics[width=\linewidth]{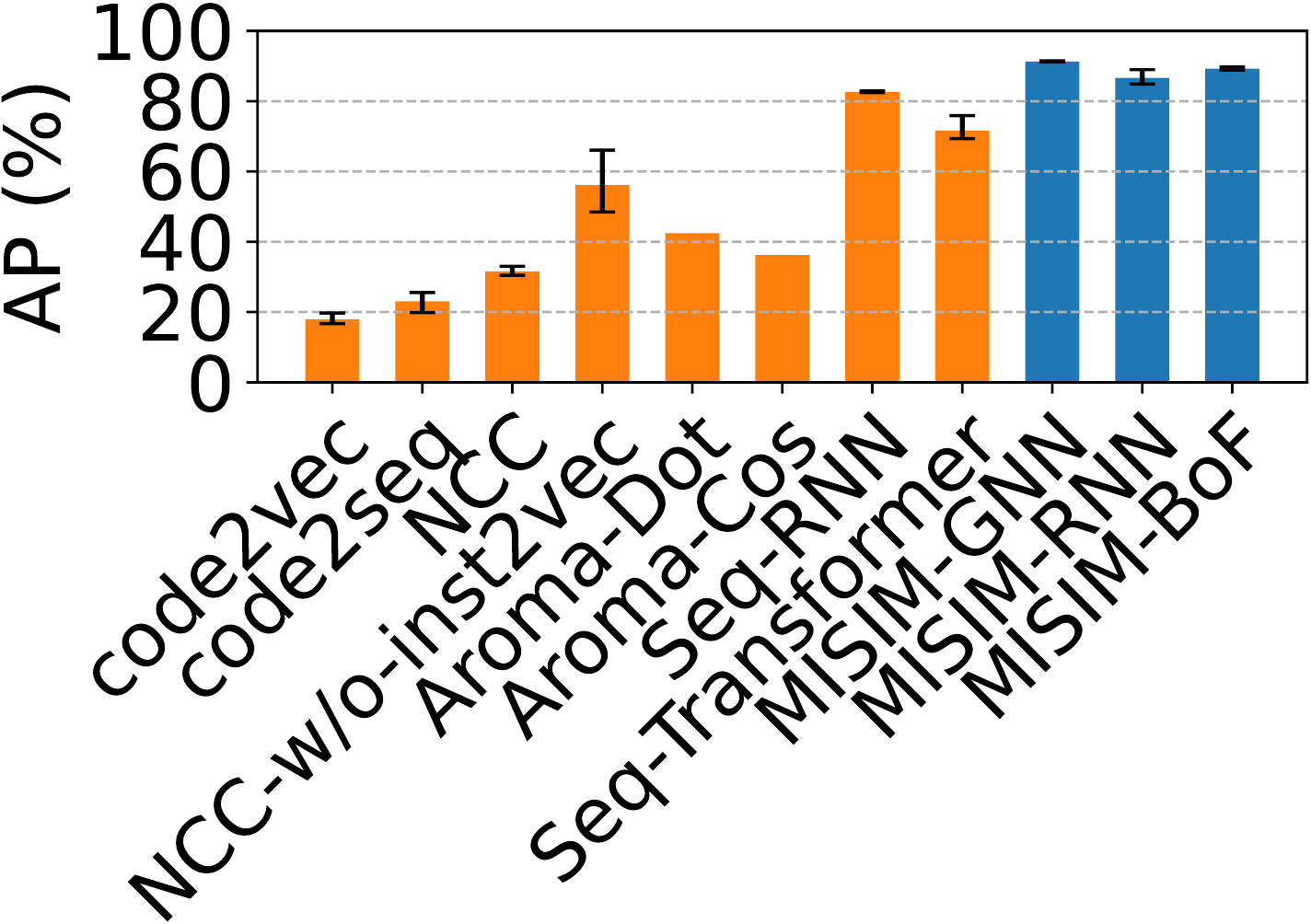}
\vspace{-15pt}
\caption{AP on GCJ.}
\end{subfigure}
\hfill
\begin{subfigure}[t]{0.24\linewidth}
\includegraphics[width=\linewidth]{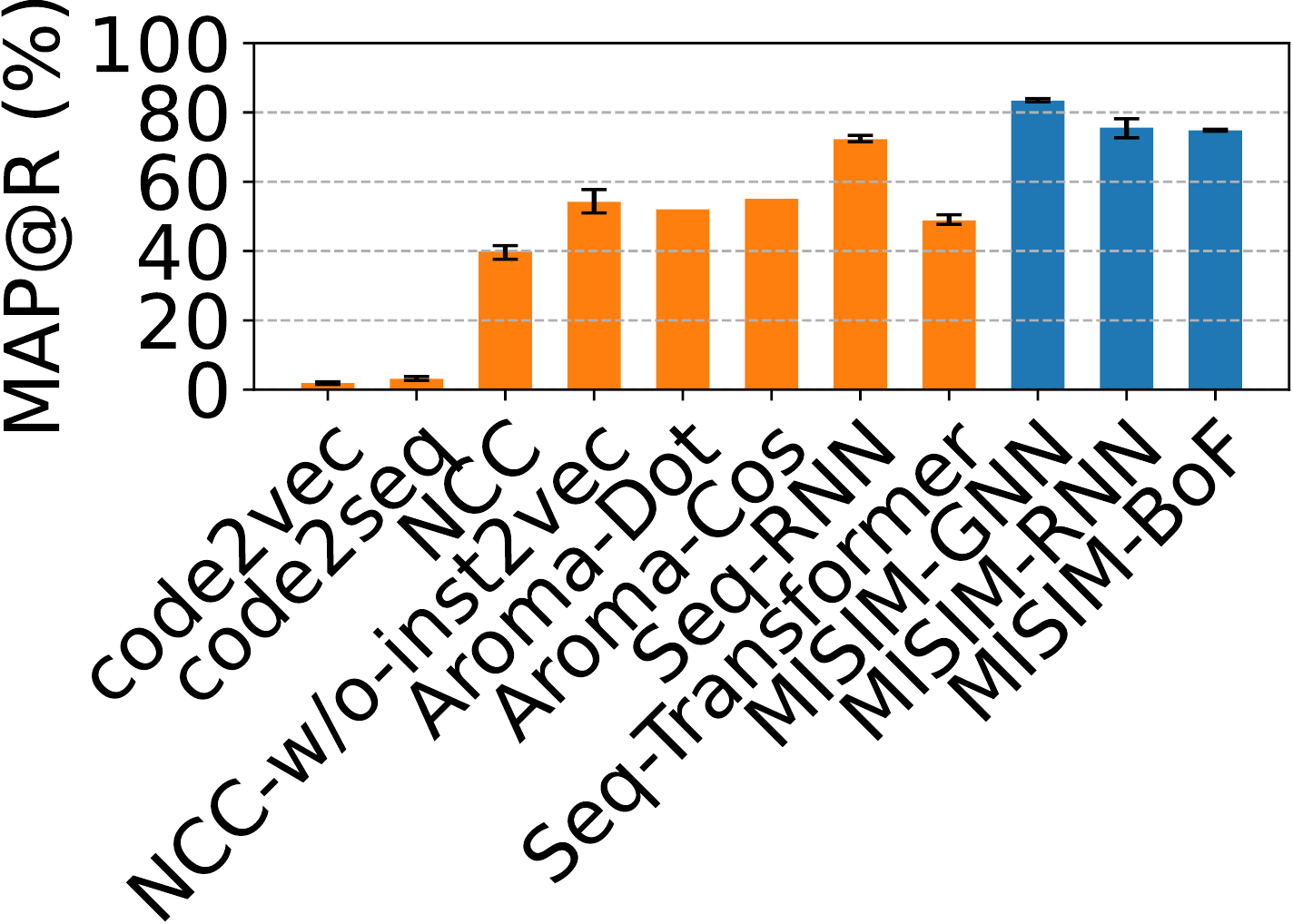}
\vspace{-15pt}
\caption{MAP@R on POJ-104.}
\end{subfigure}
\hfill
\begin{subfigure}[t]{0.24\linewidth}
\includegraphics[width=\linewidth]{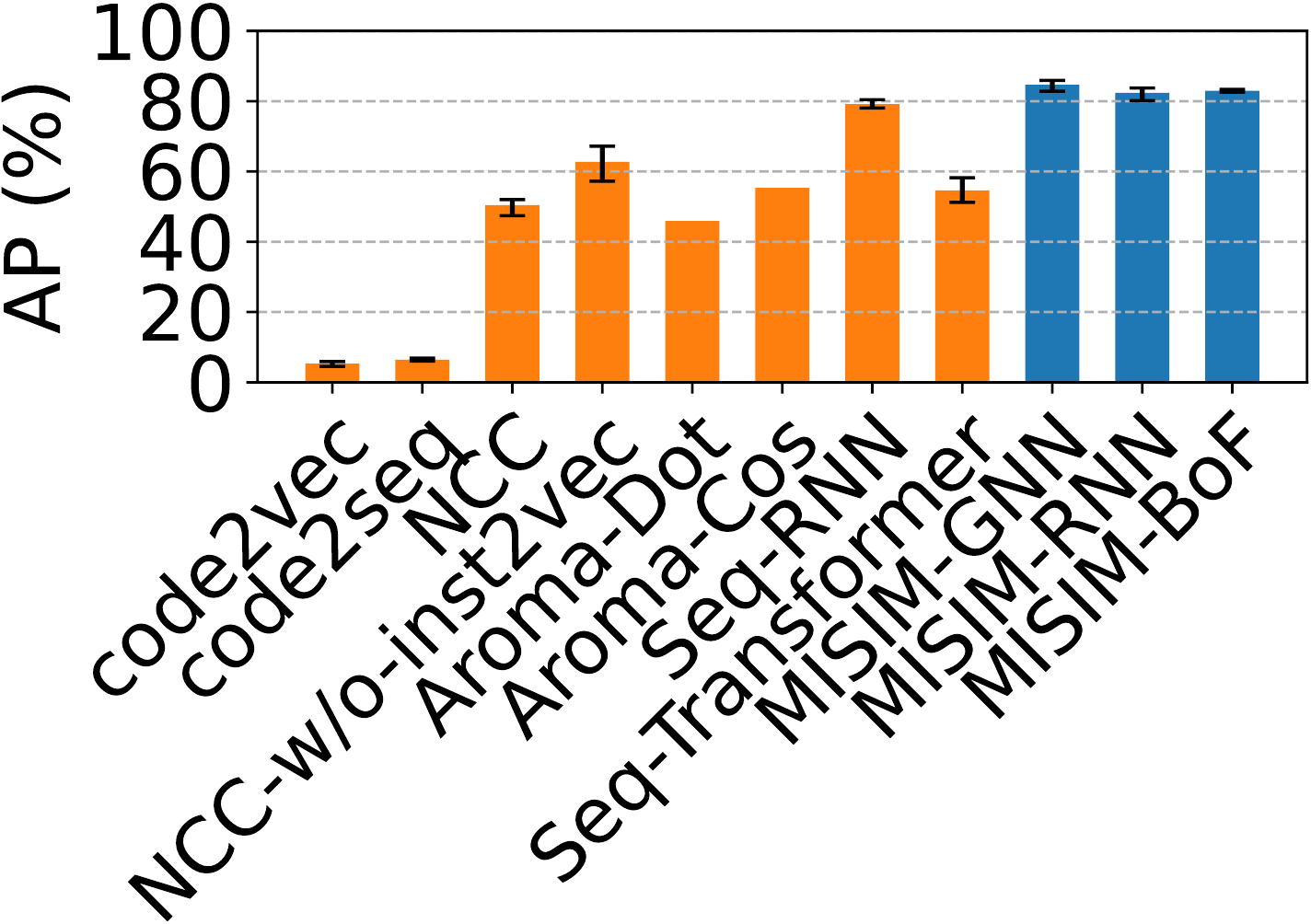}
\vspace{-15pt}
\caption{AP on POJ-104.}
\end{subfigure}
\vspace{-5pt}
\caption{Summarized accuracy results on the test sets for {\color{pltorange} code2vec, code2seq, NCC, Aroma, token sequence models}, and {\color{pltblue} MISIM} (avg over 3 runs; min/max values for error bars.)}
\vspace{-5pt}
\label{fig:generalized_result}
\end{figure*}

\begin{figure}[t]
    \centering
    \begin{subfigure}[t]{0.48\linewidth}
        \begin{subfigure}[t]{0.48\linewidth}
        \includegraphics[width=\linewidth]{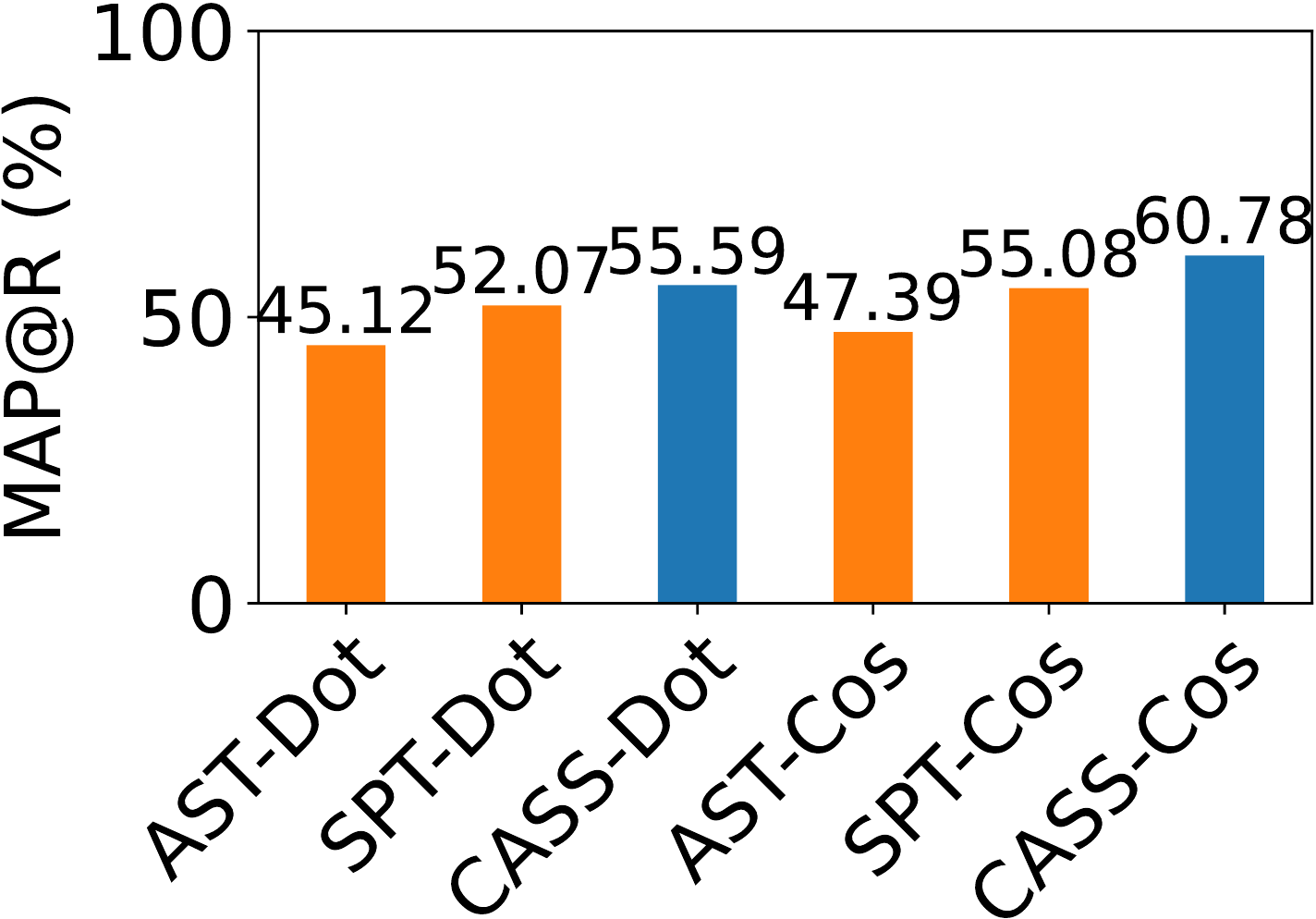}
        \vspace{-15pt}
        \end{subfigure}
        \hfill
        \begin{subfigure}[t]{0.48\linewidth}
        \includegraphics[width=\linewidth]{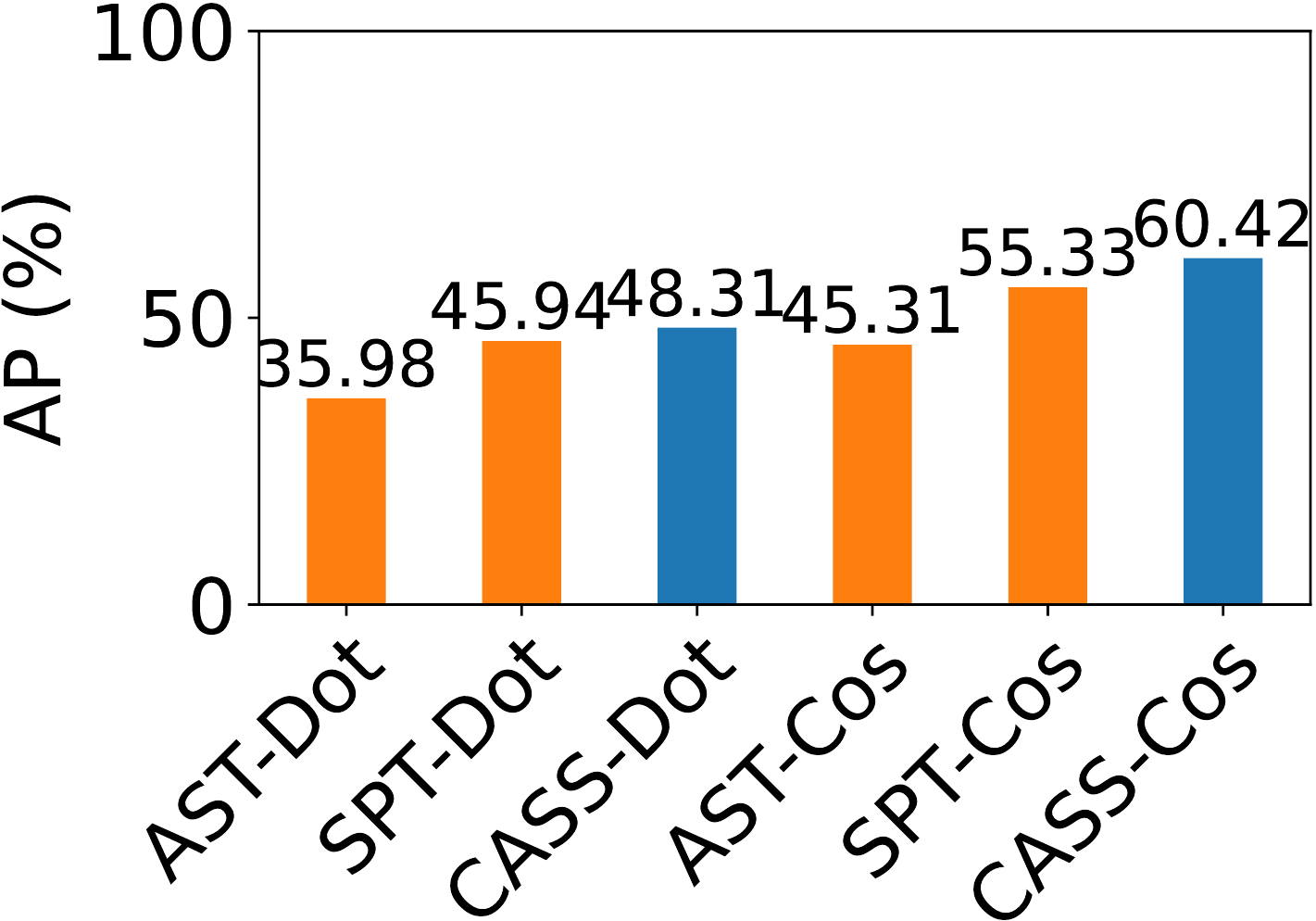}
        \vspace{-15pt}
        \end{subfigure}
    \caption{AST vs SPT vs. CASS (2-1-3-1-1)}
    \label{fig:ast_vs_spt_vs_cass_no_neural_backend}
    \end{subfigure}
    \hfill
    \begin{subfigure}[t]{0.48\linewidth}
        \begin{subfigure}[t]{0.48\linewidth}
        \includegraphics[width=\linewidth]{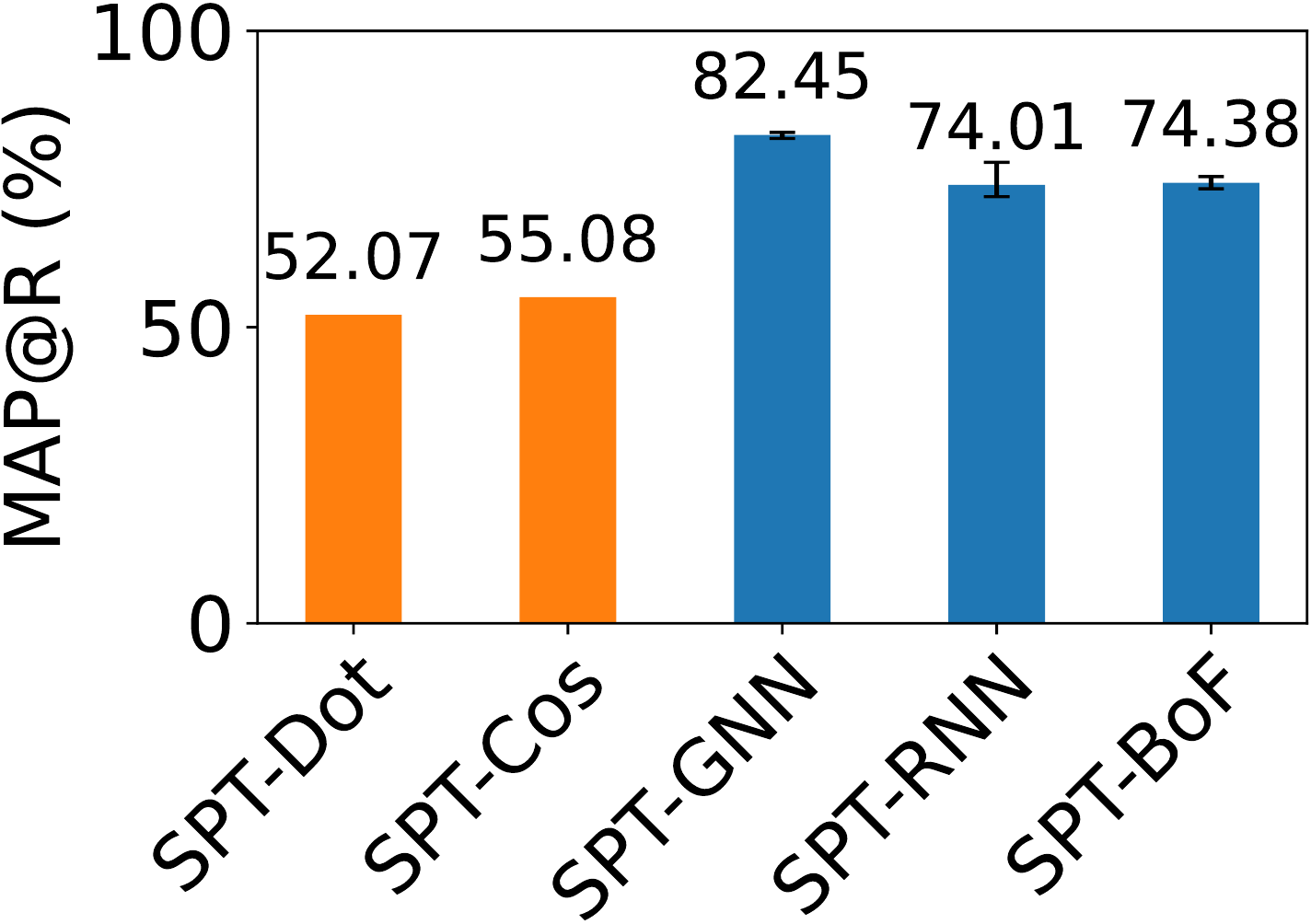}
        \vspace{-15pt}
        \end{subfigure}
        \hfill
        \begin{subfigure}[t]{0.48\linewidth}
        \includegraphics[width=\linewidth]{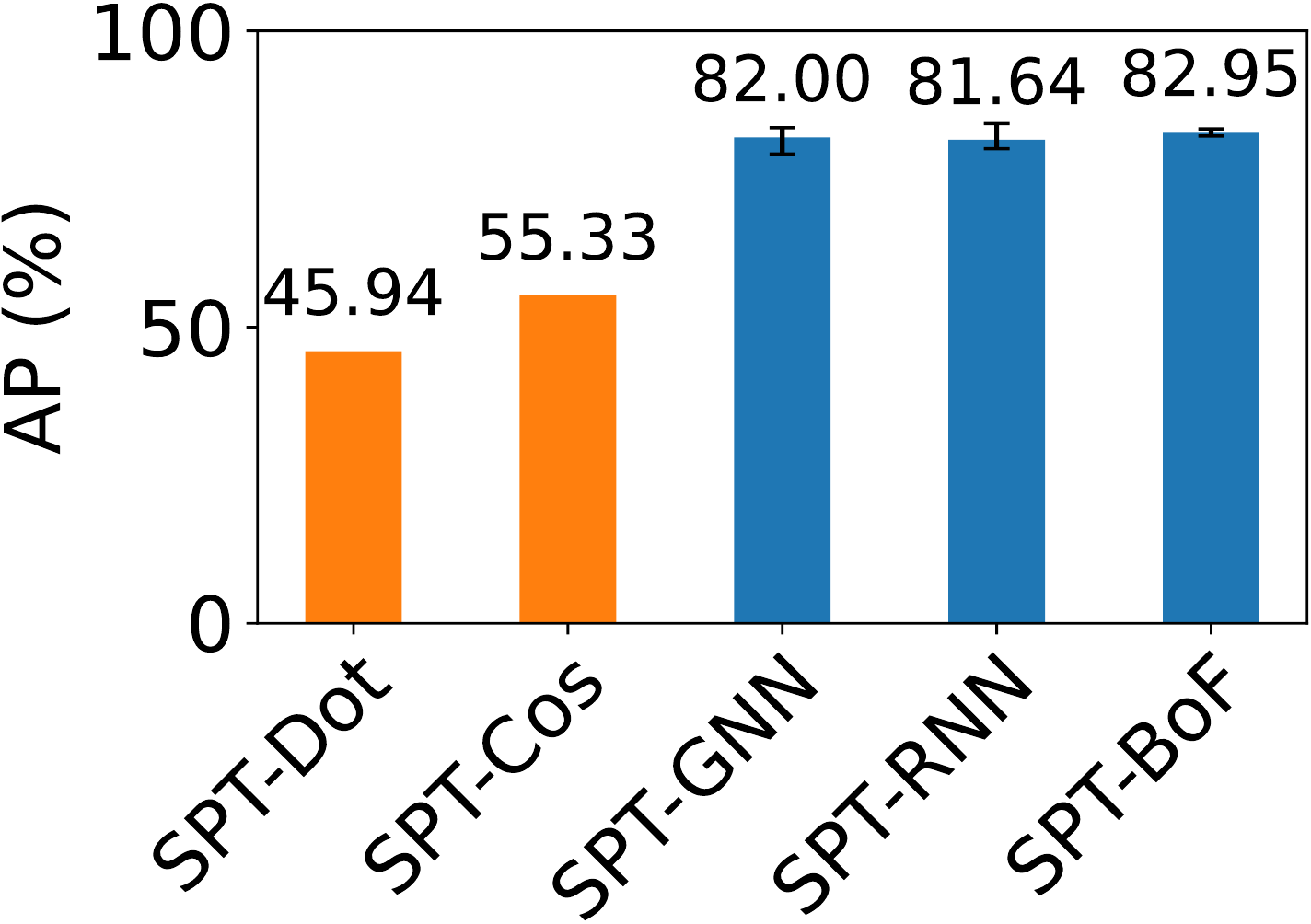}
        \vspace{-15pt}
        \end{subfigure}
     \caption{SPT vs. SPT + neural backends}
     \label{fig:spt_with_neural_backend}
     \end{subfigure}
\vspace{-5pt}
\caption{Figures showing effects of CASS configurability and neural backend separately on POJ-104.}
\end{figure}

\textbf{Configuration Identifier.}
In the following sections, we refer to a configuration of CASS by its unique identifier (ID). A configuration ID is formatted as A-B-C-D-E. Each of the five letters corresponds to a configuration type in the second column of Table~\ref{tab:configurations}, and will be replaced by an option number specified in the third column of the table. Configuration 0-0-0-0-0 corresponds to Aroma's SPT.

\subsection{End-to-End Results}

Figure~\ref{fig:generalized_result} shows the accuracy of code similarity systems compared in our experiments.\footnote{A table for these results can be found in Appendix~\ref{appendix:results_in_table_form}.} The blue bars show the results of the \system{} system variants trained using CASS configuration 2-1-3-1-1. The orange bars show the results of code2vec, code2seq, NCC, Aroma, and the two token sequence models. We observe that \system{}-GNN results in the best performance for MAP@R, yielding 1.08$\times$ to 43.89$\times$ improvements over the other systems. In summary, \system{} system performs better than other systems on both the metrics.

\begin{wraptable}{r}{150pt}
\vspace{-12pt}
\setlength\tabcolsep{10pt}
\caption{\footnotesize{Similarity MAP@R score from CodeNet (credit: ~\citep{puri:2021:arxiv}).}}
\vspace{-5pt}
\label{tab:codenet_data}
\centering
\scriptsize
\begin{tabular}{|c|c|c|}
\hline
      & C++1000 & C++1400 \\ \hline
Aroma & 0.17    & 0.15    \\ 
MISIM & 0.75    & 0.75    \\ \hline
\end{tabular}
\vspace{-10pt}
\end{wraptable}

\textbf{Additional Results.}
Table~\ref{tab:codenet_data} shows additional experimental results of Aroma and \system{} gathered by the authors of CodeNet~\citep{puri:2021:arxiv}. In their experiment, they evaluated Aroma and \system{} on code semantics similarity problem using their proposed C++1000 and C++1400 datasets. These two datasets are similar to the POJ-104 dataset. The C++1000 dataset consists of 1000 classes and 500,000 programs, and the C++1400 dataset consists of 1400 classes and 420,000 programs.\footnote{IBM/MIT's CodeNet was publicly available a week prior to the NeurIPS submissions. We only include their results on MISIM and Aroma. If accepted, we will perform a comprehensive study for camera-ready submission.}


\subsection{Specialized Experiment: Ablation Study}

To understand the effectiveness of CASS's configurability and neural backends separately, we performed two additional experiments. In the first experiment, we compared AST and Aroma's SPT representation (CASS config 0-0-0-0-0) with CASS's 2-1-3-1-1 configuration and excluded neural backends. We replaced neural backends with dot product and cosine similarity, and obtained MAP@R and AP scores on POJ-104 dataset. Figure~\ref{fig:ast_vs_spt_vs_cass_no_neural_backend} shows the results. To summarize, CASS's configuration delivers ~3-5\% better accuracy over SPT and ~10-13\% better accuracy over AST. In the second experiment, we added {\system}'s three neural backends to Aroma (Figure~\ref{fig:spt_with_neural_backend}) and found that Aroma's accuracy improves by 22-27\%. In essence, both CASS's configurability and the neural backends independently help in improving {\system}'s accuracy over existing systems.

\subsection{Specialized Experiment: CASS Configurations}

We provide early an abbreviated investigation indicating that no CASS configuration is invariably best for all code. We ran experiments that train \system{}-GNN models with two CASS configurations ($C{_1}$: 0-0-0-0-0 and $C{_2}$: 2-2-3-2-1) on randomly sampled sub-training sets and compared their accuracy. Table~\ref{tab:specialized_result} shows the results from four selected sub-training sets, named $T{_A}$, $T{_B}$, $T{_C}$, and $T{_D}$ from POJ-104. When trained on $T{_A}$ or $T{_B}$, the system using
configuration $C{_2}$ performs better than that using $C{_1}$ in both the accuracy metrics. However, for $T{_C}$ or $T{_D}$, the results are inverted.

\begin{wraptable}{r}{200pt}
\vspace{-10pt}
\caption{Test accuracy (avg, min, max) of \system-GNN trained on different training subsets}
\vspace{-5pt}
\centering
\scriptsize
\begin{tabular}{cccc}
\toprule
Subset & Config & MAP@R (\%) & AP (\%)  \\ \midrule
\multirow{2}{*}{$T{_A}$}   & $C{_1}$ & 69.78 (-0.42/+0.21)          & 76.39 (-1.68/+1.51) \\
            & $C{_2}$  & \textbf{71.99 (-0.26/+0.45)} & \textbf{79.89 (-1.20/+0.71)} \\ \midrule
\multirow{2}{*}{$T{_B}$}   & $C{_1}$ & 63.45 (-1.58/+1.92)  & 68.58 (-2.51/+2.85)   \\
            & $C{_2}$  & \textbf{67.40 (-1.85/+1.23)} & \textbf{69.86 (-3.34/+1.79)} \\ \midrule
\multirow{2}{*}{$T{_C}$}   & $C{_1}$  & \textbf{63.53 (-1.08/+1.53)} & \textbf{72.47 (-0.95/+1.24)} \\
            & $C{_2}$  & 61.23 (-2.04/+1.57)   & 69.83 (-1.03/+1.60)\\ \midrule
\multirow{2}{*}{$T{_D}$}   & $C{_1}$ & \textbf{61.78 (-0.46/+0.47)} & \textbf{66.86 (-2.31/+2.81)} \\
            & $C{_2}$  & 60.86 (-1.59/+0.90) & 63.86 (-3.06/+3.43) \\ \bottomrule
\end{tabular}
\label{tab:specialized_result}
\vspace{-5pt}
\end{wraptable}

We compared the semantics features of $T{_A} \cap T{_B}$ to $T{_C} \cap T{_D}$. We observed CASS-defined semantically salient features (e.g., global variables) that $C{_2}$ had been customized to extract, occurred less frequently in $T{_A} \cap T{_B}$. For the POJ-104 dataset, when global variables are used more frequently, they are more likely to have consistent meaning across different programs. Abstracting them away as $C_2$ does for $T{_C}, T{_D}$, leads to a loss in semantic information salient to code similarity. Conversely, when global variables are not frequently used, there is an increased likelihood that the semantics they extract are program-specific. Retaining their names in a CASS, may increase syntactic noise, reducing model performance. $C{_2}$ eliminates them for $T{_A}, T{_B}$, and has improved accuracy.



\section{Related Work} \label{sect:related}

There is a body of work on code comprehension not directly intended for code semantics similarity, but may provide value to it. Some researchers have studied applying machine learning to learn from the AST for completing various tasks on code~\citep{alon:2019:iclr, chen:2017:corr, hu:2018:icpc, li:2018:ijcai, mou:2016:aaai}. \citet{odena:2020:iclr} and \citet{odena:2021:iclr} represent a program as property signatures inferred from input-output pairs, which may be used to improve program synthesizers, amongst other things. There has also been work exploring graph representations. For bug detection and code generation, \citet{allamanis:2018:iclr} and \citet{brockschmidt:2019:iclr} represent a program as a graph with a backbone AST and additional edges representing lexical ordering and semantic relations between the nodes. \citet{dinella:2020:iclr} also use an AST-backboned graph representation of programs to learn bug fixing through graph transformation. \citet{hellendoorn:2020:iclr} introduce a simplified graph containing only AST leaf nodes for program repair. \citet{wei:2020:iclr} extract type dependency graphs from JavaScript programs for probabilistic type inference.

\section{Conclusion} \label{sect:conclusion}

This paper presented \system{}, a code semantics similarity system. \system{} has two core novelties. First, it uses the \emph{context-aware semantics structure} (CASS) designed to lift semantic meaning from code syntax. Second, it provides a neural-based code semantics similarity scoring algorithm for learning semantics similarity scoring using CASS. \citet{puri:2021:arxiv} and our experimental evaluation showed that \system{} outperforms four state-of-the-art code semantics similarity systems and two hand-optimized models. We also provided an anecdotal analysis illustrating that there may not be one universally optimal CASS configuration. An open research question for \system{} is in how to automatically derive the proper configuration of its various components for a given code corpus, specifically the CASS and neural scoring algorithms, which we plan to explore in future work.

\raggedbottom 

\bibliographystyle{plainnat}
\bibliography{main}

\clearpage


\clearpage

\appendix
\section{Context-Aware Semantics Structure Details}
\label{appendix:cass}

The following is the formal definition of CASS.

\begin{definition}[Context-aware semantics structure (CASS)]
\label{cass-def}
A CASS consists of one or more CASS trees and an optional global attributes
table (GAT). A CASS tree, $T$, is a collection of nodes, $V = \{v_{1},v_{2},
\dots, v_{|V|}\}$, and edges, $E = \{e_{1},e_{2}, \dots, e_{|E|}\}$, denoted as
$T = (V,E)$. Each edge is directed from a parent node, $v_{p}$ to a child node,
$v_{c}$, or $e_{k}=(v_{p},v_{c})$ where $e_{k} \in E$ and $v_{p},v_{c} \in V$.
The root node, $v_{r}$, of the tree signifies the beginning of the code snippet
and has no parent node, i.e., $\nexists v_p, (v_p,v_r)\in E$. A child node is
either an internal node or a leaf node. An internal node has at least one child
node while a leaf node has no child nodes. A CASS tree can be empty, in which it
has no nodes. The CASS GAT contains exactly one entry per unique function
definition in the code snippet. A GAT entry includes the input and output
cardinality values for the corresponding function.
\end{definition}

\begin{definition}[Node labels]
\label{node-label}
Every CASS node has an associated \textit{label}, $l_{v}$. During the
construction of a CASS tree, the program tokens at each node, $t_{v}$ are mapped
to its corresponding label or $l_{v} = f(t_{v})$.  This is depicted with an
expression grammar for node labels and the function mapping tokens to labels
below.\footnote{Note: the expression grammar we provide is non-exhaustive due to
space limitations. The complete set of standard C/C++ tokens or binary and unary
operators is collectively  denoted in shorthand as `...'.} 
\end{definition}

\vspace{-8pt}

\grammarindent=110pt
\grammarparsep=2pt
\begin{grammar}
<bin-op> ::= `+' | `-' | `*' | `/' | ...

<unary-op> ::= `++' | `-\/-' | ...

<leaf-node-label> ::= "LITERAL" | "IDENT" | `#VAR' | `#GVAR' | `#EXFUNC' | `#LIT' | ... 

<exp> ::=  `\$' | `\$'  <bin-op> `\$' | <unary-op> `\$' | ...

<internal-node-label> ::= `for' `(' <exp> `;' <exp> `;' <exp> `)' <exp> `;'
    \alt `int' <exp> `;'
    \alt `return' <exp> `;'
    \alt <exp>
    \alt ...
\end{grammar}

\vspace{-18pt}

\begin{align*}
l_{v} = f(t_{v}) = 
\begin{cases}
    \langle \textit{leaf-node-label} \rangle & \text{if $v$ is a leaf node} \\
    \langle \textit{internal-node-label} \rangle & \text{otherwise}
\end{cases}
\end{align*}

\begin{definition}[Node prefix label]
\label{prefix}
A node prefix label is a string prefixed to a node label. A node prefix label may or may not be present.
\end{definition}

\subsection{Discussion}
\label{subsec:intuit}

We believe there is no silver bullet solution for code similarity for all
programs and programming languages. Based on this belief, a key intuition of
CASS's design is to provide a structure that is semantically rich based on
structure, with inspiration from Aroma's SPT, while simultaneously providing a
range of customizable parameters to accommodate a wide variety of scenarios.
CASS's language-agnostic and language-specific configurations and their
associated options serve for exploration of a series of tree variants, each
differing in their granularity of detail of abstractions. 

For instance, the \textit{compound statements} configuration provides three
levels of abstraction. Option 0 is Aroma's baseline configuration and is the
finest level of abstraction, as it featurizes the number of constituents in a
compound statement node. Option 2 reduces compound statements to a single token
and represents a slightly higher level of abstraction. Option 1 eliminates all
features related to compound statements and is the coarsest level of
abstraction. The same trend applies to the \textit{global variables} and
\textit{global functions} configurations. It is our belief, based on early
evidence, that the appropriate level of abstraction in CASS is likely based on
many factors such as \emph{(i)} code similarity purpose, \emph{(ii)} programming
language expressiveness, and \emph{(iii)} application domain.

Aroma's original SPT seems to work well for a common code base where global
variables have consistent semantics and global functions are standard API calls
also with consistent semantics (e.g., a single code-base). However, for cases
outside of such spaces, some question about applicability arise. For example,
assumptions about consistent semantics for global variables and functions may
not hold in cases of non-common code-bases or non-standardized global function
names~\citep{wulf:1973:sigplan, gellenbeck:1991:ablex, feitelson:2020:ieee}.
Having the capability to differentiate between these cases, and others, is a key
motivation for CASS.

We do not believe that CASS's current structure is exhaustive. With this in
mind, we have designed CASS to be extensible, enabling a seamless mechanism to
add new configurations and options. Our intention with this paper is to
present initial findings in exploring CASS's structure. Based on our early
experimental analysis, presented in Section~\ref{sect:appendix-experiments-cass-configurations},
CASS seems to be a promising research direction for code similarity.

\paragraph{An Important Weakness.}
\label{ssect:weakness}

While CASS provides added flexibility over SPT, such flexibility may be misused.
With CASS, system developers are free to add or remove as much syntactic
differentiation detail they choose for a given language or given code body. Such
overspecification (or  underspecification), may result in syntactic overload (or
underload) which may cause reduced code similarity accuracy over the original
SPT design, as we illustrate in Section~\ref{sect:appendix-experiments-cass-configurations}.

\section{MISIM Models}
\label{appendix:models}

In this section, we describe the models evaluated in our experiments other than \system{}-GNN, and discuss the details of the experimental procedure.

\subsection{\system{}-BoF}

The \system{}-BoF model takes a set of manual features extracted from a CASS as its input. The features include the ones extracted from CASS trees, using the same procedure described in Aroma~\citep{luan:2019:oopsla}, as well as the entries in CASS GATs. The \system{}-BoF model is specified as below:
\begin{align*}
\vec{c} = \mathrm{FC} \left(
\mathrm{AvgPool} \left(\left\{ \vec{e}_x \;\middle|\; x \in S \right\}\right)
\right)
,
\end{align*}
where $S$ is the feature set of the input program and $\vec{e}_x$ is the embedding vector of feature $x$.
The output code vector is computed by performing average pooling on the feature embeddings and projecting its result into the code vector space with a fully connected layer.


\subsection{\system{}-RNN}

The input to the \system{}-RNN model is a serialized representation of a CASS.
Each CASS tree, representing a function in the program, is converted to a
sequence using the technique proposed in~\citep{hu:2018:icpc}. The GAT entry
associated with a CASS tree is both prepended and appended to the tree's
sequence, forming the sequence of the corresponding function.
The model architecture can be expressed as:
\begin{gather*}
\vec{h}_f = \mathrm{biGRU} \left( \Bar{\vec{e}}_f \right)
,\\
\vec{c} = \mathrm{FC} \left( \left[
\mathrm{AvgPool} \left(\left\{ \vec{h}_f \;\middle|\; f \in F \right\}\right) ;
\mathrm{MaxPool} \left(\left\{ \vec{h}_f \;\middle|\; f \in F \right\}\right)
\right] \right)
,
\end{gather*}
where $F$ is the set of functions in the input program and $\Bar{\vec{e}}_f$ is the sequence of embedding vectors for the serialized CASS of function $f$.
Each function's sequence first has its tokens embedded,
and then gets summarized to a function-level vector by a bidirectional GRU
layer~\citep{cho:2014:emnlp}. The code vector for the entire program is
subsequently computed by performing average and max pooling on the function-level
vectors, concatenating the resulting vectors, and passing it
through a fully connected layer.

\section{Experimental Details}
\label{appendix:experimental_details}

\subsection{Training Procedure and Hyperparameters}
\label{appendix:training}

We use the AdamW optimizer~\citep{loshchilov:2019:iclr} with a learning rate of $10^{-3}$ for all the models except Seq-Transformer, for which a learning rate of $10^{-4}$ is used to stabilize training. The training runs for 100 epochs, each containing 1,000 iterations, and the model that gives the best validation accuracy is used for testing.\footnote{We have observed that the validation accuracy stops to increase before the 100\textsuperscript{th} epoch in all experiments.} The hyperparameters used for the Circle loss are $\gamma=80$ and $m=0.4$. For all of our \system{} models, we use 128-dimensional embedding vectors, hidden states, and code vectors. We also apply dropout with a probability of 0.5 to the embedding vectors. To handle rare or unknown tokens, a token that appears less than 5 times in the training set is replaced with a special \texttt{UNKNOWN} token.

\subsection{Modifications to code2vec, code2seq, NCC, and Aroma}
\label{appendix:modification_to_others}

To compare with code2vec, code2seq, NCC, and Aroma, we adapt them to our experimental
setting in the following ways. The original code2vec takes a function as an
input, extracts its AST paths to form the input to its neural network, and
trains the network using the function name prediction task. In our experiments,
we feed the AST paths from all function(s) in a program into the neural network
and train it using the metric learning task described in
Section~\ref{sec:misim_training}.
We make similar adaptions to code2seq by combining AST paths from the whole program as one input sample.
Additionally, we replace the sequence decoder of code2vec with an attention-based path aggregator used in code2vec.
NCC contains a pre-training phase, named
\texttt{inst2vec}, on a large code corpus for generating instruction embeddings,
and a subsequent phase that trains an RNN for a downstream task using the
pre-trained embeddings. We train the downstream RNN model on our metric learning
task in two ways. The first uses the pre-trained embeddings (labeled as NCC in
our results). The second trains the embeddings from scratch on our task in an
end-to-end fashion (labeled as NCC-w/o-inst2vec). For both code2vec and NCC,
we use the same model architectures and embedding/hidden sizes suggested in
their papers and open-sourced implementations. The dimension of their output
vectors (i.e., code vectors) is set to the same as our \system{} models. Aroma
extracts manual features from the code and computes the similarity score of two
programs by taking the dot product of their binary feature vectors. We
experiment with both its original scoring mechanism (labeled as Aroma-Dot) and a
variant that uses the cosine similarity (labeled as Aroma-Cos).

\subsection{Token Sequence Models}
\label{appendix:seq_models}

Here we provide details of the two token sequence model (Seq-RNN and Seq-Transformer) used in our experiments. 

The input to both models is tokenized source code. Identifiers and non-identifier tokens are embedded differently. For identifiers, we first split their text values into subtokens according to camel case and snake case patterns. Then we embed each subtoken into a trainable vector and generate an identifier's embedding by taking the sum of its subtokens' embedding vectors. For non-identifier tokens, we assign a trainable embedding vector to each unique token. All embedding vectors are 128-dimensional.

Seq-RNN employs a stacked two-layer bidirectional GRU that operates on token embedding sequences. Each direction in a layer has a 128-dimensional hidden state, and a dropout of rate 0.1 is applied between the two layers. The last hidden states in the second layer in both directions are concatenated and projected into a 128-dimensional code vector using a linear layer. This code vector is then used to compute cosine similarities between programs.

Seq-Transformer employs a six-layer Transformer encoder with a learned positional embedding. The dimensionality of the Transformer is 128, and the feed-forward layers' hidden states are 512 dimensional. A dropout of rate 0.1 is applied inside each Transformer encoder layer. The length of the input token sequence is capped to 512. A special token is prepended to each sequence, and the output at the location of this token after the last encoder layer is passed into a linear layer to generate a 128-dimensional code vector for computing cosine similarities.

\subsection{Evaluation Metrics}
\label{appendix:evaluation_metrics}

MAP@R measures how accurately a model can retrieve similar (or relevant) items
from a database given a query. MAP@R rewards a ranking system (e.g., a search
engine, a code recommendation engine, etc.) for correctly ranking relevant items
with an order where more relevant items are ranked higher than less relevant
items. It is defined as the mean of average precision scores, each of which is
evaluated for retrieving R most similar samples given a query. In our case, the
set of queries is the set of all test programs. For a program, R is the number
of other programs in the same class (i.e., a POJ-104 problem). MAP@R is applied
to both validation and testing. We use AP to measure the performance
in a binary classification setting, in which the models are viewed as binary
classifiers that determine whether a pair of programs are similar by comparing
their similarity score with a threshold. AP is only used for testing.
They are computed from the similarity scores of all program pairs in the test
set, as well as their pair-wise labels. For the systems that require training
(i.e., systems with ML learned similarity scoring), we train and evaluate them
three times with different random seeds.

\subsection{\system{} Accuracy Results (in tabular form)}
\label{appendix:results_in_table_form}

Table~\ref{tab:generalized_result} shows the results of \system{} in comparison to
other systems. Same results are presented in the graphical form in Figure~\ref{fig:generalized_result}.

\begin{table*}[htbp]
\setlength\tabcolsep{4pt}
\caption{Code similarity system accuracy. Results are shown as the average and
min/max values, relative to the average, over 3 runs. We had to make a few
modifications to adapt code2vec, code2seq, NCC and Aroma to our experimental settings.
Please refer to Appendix~\ref{appendix:modification_to_others} for details.
}
\label{tab:generalized_result}
\centering
\scriptsize
\begin{tabular}{@{\hspace{1pt}}lcccc@{\hspace{1pt}}}
\toprule
\multirow{2}{*}{Method} &
  \multicolumn{2}{c}{GCJ} &
  \multicolumn{2}{c}{POJ-104} \\
\cmidrule(l{3pt}r{3pt}){2-3} \cmidrule(l{3pt}){4-5}
 &
  \multicolumn{1}{c}{MAP@R (\%)} &
  \multicolumn{1}{c}{AP (\%)} &
  MAP@R (\%) &
  AP (\%) \\
\cmidrule(r{3pt}){1-1} \cmidrule(l{3pt}r{3pt}){2-3} \cmidrule(l{3pt}){4-5}
code2vec &
  7.76 (-0.79/+0.88) &
  17.95 (-1.24/+1.76) &
  1.90 (-0.43/+0.38) &
  5.30 (-0.80/+0.60) \\
code2seq &
  11.67 (-1.98/+1.73) &
  23.09 (-3.24/+2.49) &
  3.12 (-0.45/+0.67) &
  6.43 (-0.37/+0.48) \\
NCC &
  17.26 (-1.11/+0.57) &
  31.56 (-1.11/+1.46) &
  39.95 (-2.29/+1.64) &
  50.42 (-2.98/+1.61) \\
NCC-w/o-inst2vec &
  34.88 (-5.72/+7.63) &
  56.12 (-7.63/+9.96) &
  54.19 (-3.18/+3.52) &
  62.75 (-5.49/+4.42) \\
Aroma-Dot &
  29.08 &
  42.47 &
  52.07 &
  45.94 \\
Aroma-Cos &
  29.67 &
  36.21 &
  55.08 &
  55.33 \\
Seq-RNN &
  69.27 (-0.52/+0.54) &
  82.66 (-0.23/+0.27) &
  72.28 (-0.76/+1.10) &
  79.19 (-1.11/+1.27) \\
Seq-Transformer &
  47.81 (-2.28/+2.90) &
  71.66 (-2.30/+4.22) &
  48.81 (-1.10/+1.66) &
  54.65 (-3.41/+3.55) \\
\cmidrule(r{3pt}){1-1} \cmidrule(l{3pt}r{3pt}){2-3} \cmidrule(l{3pt}){4-5}
\system-GNN &
  \textbf{74.87 (-0.10/+0.15)} &
  \textbf{91.32 (-0.18/+0.20)} &
  \textbf{83.39 (-0.30/+0.59)} &
  \textbf{84.69 (-1.82/+1.24)} \\
\system-RNN &
  72.50 (-3.62/+2.09) &
  86.65 (-1.76/+2.32) &
  75.61 (-2.97/+2.54) &
  82.37 (-2.20/+1.36) \\
\system-BoF &
  71.25 (-0.64/+0.42) &
  89.28 (-0.41/+0.45) &
  74.85 (-0.27/+0.31) &
  82.97 (-0.36/+0.41 \\
\bottomrule
\end{tabular}
\end{table*}

\subsection{CASS vs AST}

Some recent research on code representation uses the AST-based representation \citep{dinella:2020:iclr} or AST paths \citep{alon:2019:popl, alon:2019:iclr}. In this subsection, we explore how AST and CASS perform on the task of code semantic representation described here.

We compared the code similarity performance of ASTs and CASSes on the test set of POJ-104, as shown in table~\ref{tab:dataset_stats}, by transforming both kinds of representations into feature vectors. using the same method described in \citep{luan:2019:oopsla} and compute the similarity scores using dot or cosine similarity. For each program in the dataset, we extracted its CASS under three different configurations: 0-0-0-0-0~\footnote{Configuration 0-0-0-0-0 is the duplicate of SPT. As shown in Table~\ref{tab:ast_cass_poj104}, configuration 2-1-3-1-1 shows better accuracy than configuration 0-0-0-0-0.}, the base configuration, and 2-1-3-1-1/1-2-1-0-0, the best/worst performing configuration according to our preliminary evaluation of CASS (see Appendix~\ref{sect:appendix-experiments-cass-configurations} for details).
We also extracted the ASTs of function bodies in a program. Each syntax node in the AST is labeled by its node type, and an identifier (or literal) node also gets a single child labeled by the corresponding identifier name (or literal text).


\begin{table}[htbp]
\centering
\scriptsize
\caption{Test Accuracy for AST and CASS configurations on POJ-104.}
\begin{tabular}{lccc}
\toprule
Method         & MAP@R (\%) & AP (\%) \\ \midrule
AST-Dot        & 45.12      & 35.98      \\
AST-Cos        & 47.39      & 45.31      \\
\midrule
SPT-Dot  & 52.07      & 45.94      \\
SPT-Cos  & 55.08      & 55.33      \\
\midrule
CASS (2-1-3-1-1)-Dot & 55.59      & 48.31       \\
CASS (2-1-3-1-1)-Cos & \textbf{60.78}      & \textbf{60.42} \\
CASS (1-2-1-0-0)-Dot & 52.74      & 40.73   \\
CASS (1-2-1-0-0)-Cos & 57.99      & 54.75   \\ \bottomrule
\end{tabular}
\label{tab:ast_cass_poj104}
\end{table}

As shown in Table~\ref{tab:ast_cass_poj104}, CASS configurations show an improvement in accuracy over the AST up to $1.33 \times$ in both evaluation metrics described in Appendix~\ref{appendix:evaluation_metrics}. To better understand the performance difference, we investigated a few solutions for the same problems from the POJ-104 dataset. One of the interesting observations we found is that for the same problem, a solution may have a different naming convention for local variables than that of another solution (e.g., English vs Mandarin description of variables), but the resulting different variable names may carry the same semantic meaning. AST uses variable names in its structure, but CASS has the option to not use variable names. Thus the erasure of local variable names in CASS might help in discovering the semantic similarity between code with different variable names. This might explain some of the performance differences between AST and CASS in this experiment.

\subsection{Experimental Results of Various CASS configurations}
\label{sect:appendix-experiments-cass-configurations}

In this section, we discuss our experimental setup and analyze the performance
of CASS compared to Aroma's simplified parse tree (SPT).
In Section~\ref{subsec:setup}, we explain the dataset grouping and enumeration
for our experiments. We also discuss the metrics used to quantitatively rank the
different CASS configurations and those chosen for the evaluation of code
similarity. Section~\ref{subsec:results} demonstrates that,  a code similarity
system built using CASS \emph{(i)} has a greater frequency of improved accuracy
for the total number of problems and \emph{(ii)} is, on average, more accurate
than SPT. For completeness, we also include cases where CASS configurations
perform poorly.

\subsubsection{Experimental Setup}
\label{subsec:setup}
In this section, we describe our experimental setup. At the highest level, we compare the performance of various configurations of CASS to Aroma's SPT. The list of possible CASS configurations is shown in Table~\ref{tab:configurations}.

\paragraph{Dataset.}
The experiments use the same POJ-104 dataset introduced in Section~\ref{sect:experiments}.

\paragraph{Problem Group Selection.}
Given that POJ-104 consists of 104 unique problems and nearly 50,000 programs, depending on how we analyze the data, we might face intractability problems in both computational and combinatorial complexity. With this in mind, our initial approach is to construct 1000 sets of five unique, pseudo-randomly selected problems for code similarity analysis. Using this approach, we evaluate every configuration of CASS and Aroma's original SPT on each pair of solutions for each problem set. We then aggregate the results across all the groups to estimate their overall performance. While this approach is not exhaustive of possible combinations (in set size or set combinations), we aim for it to be a reasonable starting point. As our research with CASS matures, we plan to explore a broader variety of set sizes and a more exhaustive number of combinations.

\paragraph{Code Similarity Performance Evaluation.}
For each problem group, we exhaustively calculate code similarity scores for all unique solution pairs, including pairs constructed from the same program solution (i.e., program $A$ compared to program $A$). We use $G$ to refer to the set of groups and $g$ to indicate a particular group in $G$. We denote $|G|$ as the number of groups in $G$ (i.e. cardinality) and |g| as the number of solutions in group $g$. For $g$ = $G_i$, where $i = \{1, 2, \ldots, 1000\}$, the total unique program pairs (denoted by $g_P$) in $G_i$ is $\mathit{|g_{P}|} = \frac{1}{2}|g|(|g|+1)$.

To compute the similarity score of a solution pair, we use Aroma's approach. This includes calculating the dot product of two feature vectors (i.e., a program pair), each of which is generated from a CASS or SPT structure. The larger the magnitude of the dot product, the greater the similarity.

We evaluate the quality of the recommendation based on \emph{average precision}. \emph{Precision} is the ratio of true positives to the sum of true positives and false positives. Here, true positives denote solution pairs correctly classified as similar and false positives refer to solution pairs incorrectly classified as similar. \emph{Recall} is the ratio of true positives to the sum of true positives and false negatives, where false negatives are solution pairs incorrectly classified as different. As we monotonically increase the threshold from the minimum value to the maximum value, precision generally increases while recall generally decreases. The \textit{average precision} (AP) summarizes the performance of a binary classifier under different thresholds for categorizing whether the solutions are from the same equivalence class (i.e., the same POJ-104 problem)~\citep{liu:2009:now}. AP is calculated using the following formula over all thresholds.

\begin{enumerate}
    \item All unique values from the $M$ similarity scores, corresponding to the solution pairs, are gathered and sorted in descending order. Let $N$ be the number of unique scores and $s_1, s_2, \ldots, s_N$ be the sorted list of such scores.
    
    \item For $i$ in $\{1, 2, \ldots, N\}$, the precision $p_i$ and recall $r_i$ for the classifier with the threshold being $s_i$ is computed.
    \item Let $r_0 = 0$. The average precision is computed as: $$AP = \sum_{i=1}^N(r_i-r_{i-1})p_i$$
\end{enumerate}

\subsubsection{Results}
\label{subsec:results}

\begin{figure*}[htbp]
\centering
\begin{subfigure}[t]{0.3\linewidth}
\includegraphics[width=\linewidth]{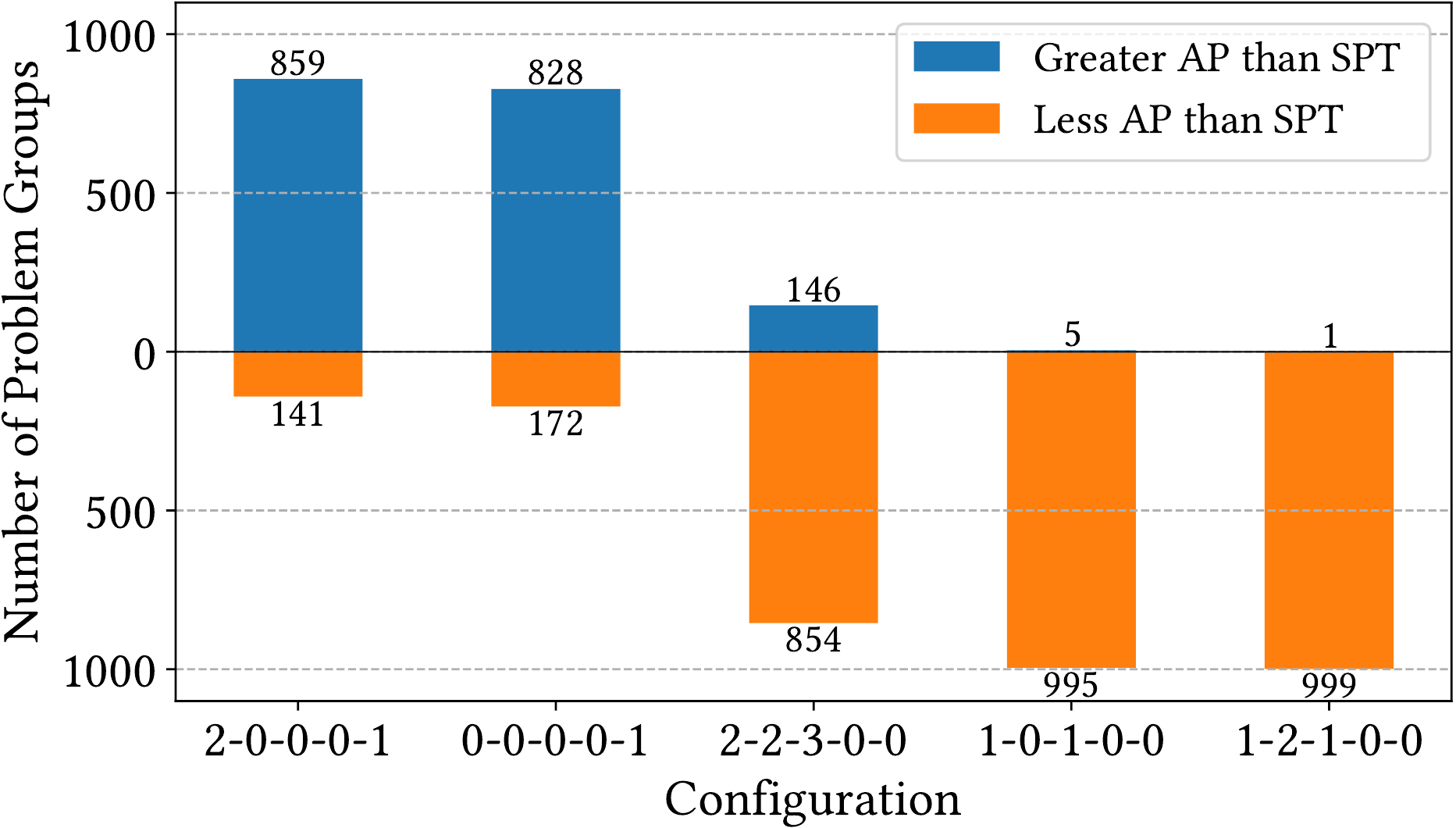}
\caption{Breakdown of the Number of Groups with AP Greater or Less than SPT.}
\label{fig:ap_gl}
\end{subfigure}
\hfill
\begin{subfigure}[t]{0.3\linewidth}
\includegraphics[width=\linewidth]{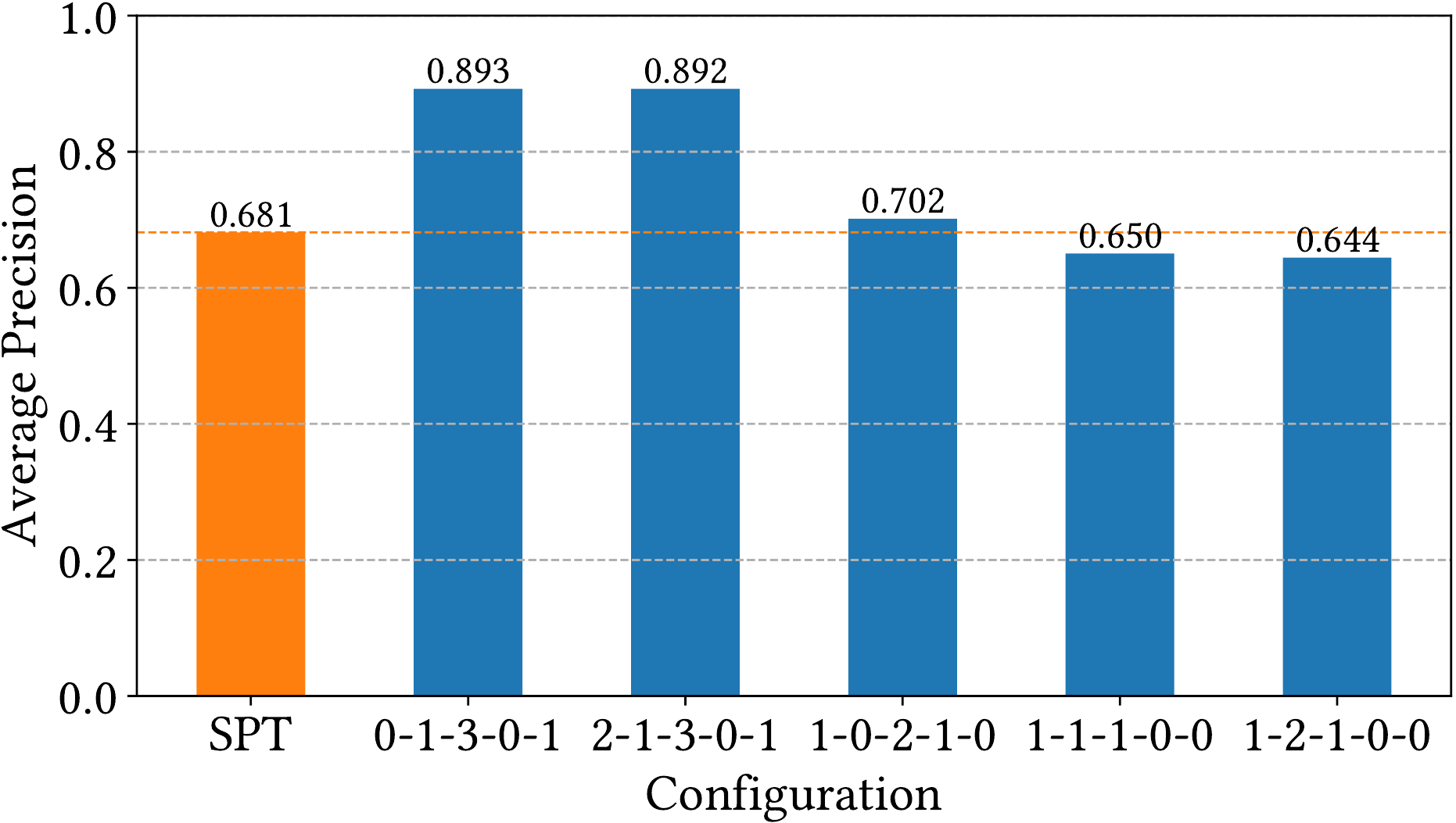}
\caption{Average Precision for the Group Containing the Best Case.}
\label{fig:best_group}
\end{subfigure}
%
\hfill
\begin{subfigure}[t]{0.3\linewidth}
\includegraphics[width=\linewidth]{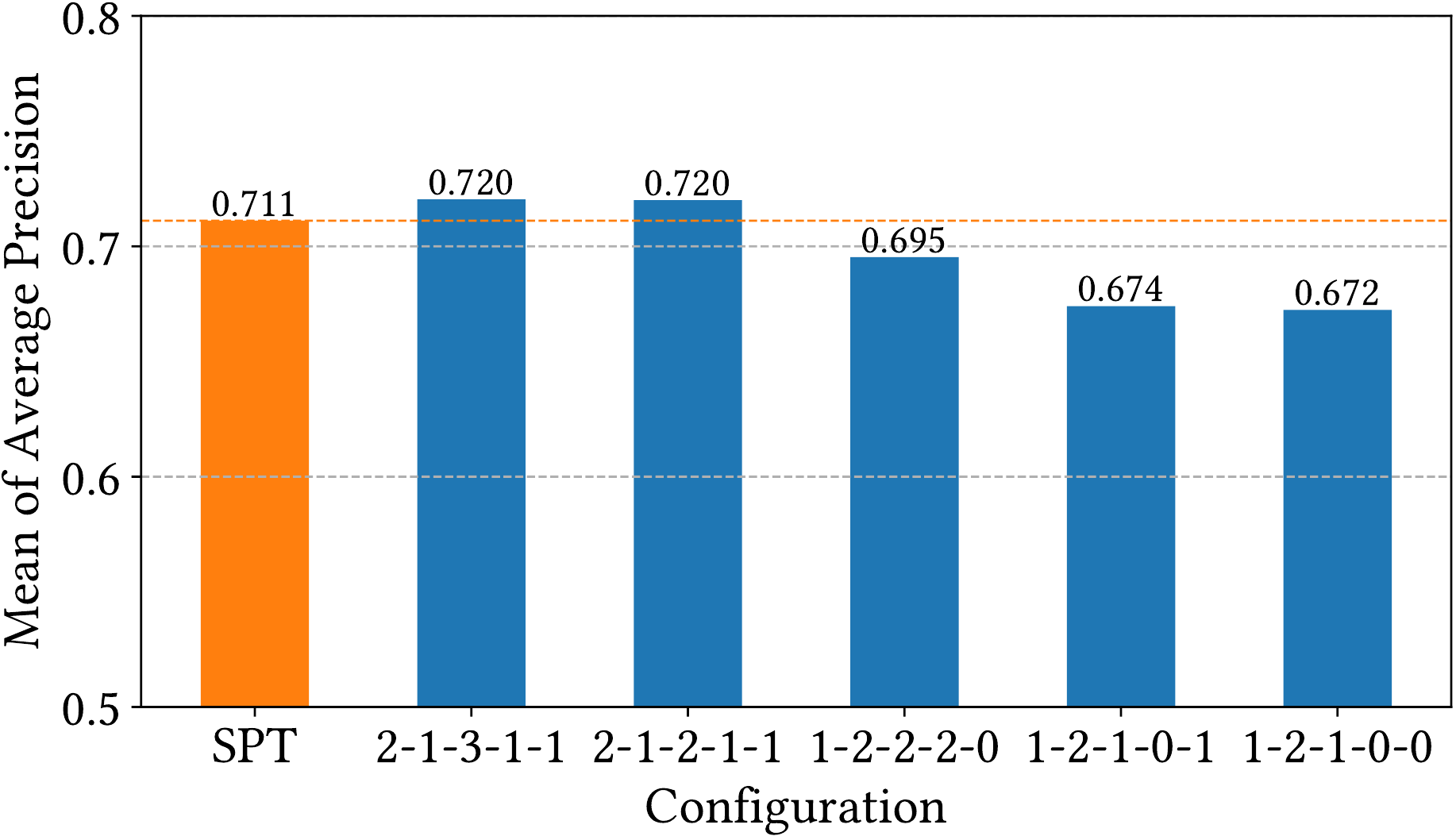}
\caption{Mean of Average Precision Over All Program Groups.}
\label{fig:ap_mean}
\end{subfigure}
\caption{Comparison of CASS and SPT. The blue bars in (a) and (b), and all the bars in (c), from left to right, correspond to the best two, the median, and the worst two CASS configurations, ranked by the metric displayed in each subfigure.}
\label{fig:ap_all}
\end{figure*}

Figure~\ref{fig:ap_gl} depicts the number of problem groups where a particular CASS variant performed better (blue) or worse (orange) than SPT. For example, the CASS configuration 2-0-0-0-1 outperformed SPT in 859 of 1000 problem groups, and underperformed in 141 problem groups. This equates to a 71.8\% accuracy improvement of CASS over SPT. Figure~\ref{fig:ap_gl} shows the two best (2-0-0-0-1 and 0-0-0-0-1), the median (2-2-3-0-0), and the two worst (1-0-1-0-0 and 1-2-1-0-0) configurations with respect to SPT. Although we have seen certain configurations that perform better than SPT, there are also configurations that perform worse. We observed that the configurations with better performance have function I/O cardinality option as 1. We also observed that the configurations with worse performance have function I/O cardinality option as 0. These observations indicate that function I/O cardinality seems to improve code similarity accuracy, at least, for the data we are considering. We speculate that these configuration results may vary based on programming language, problem domain, and other constraints.

Figure~\ref{fig:best_group} shows the group containing the problems for which CASS achieved the best performance relative to SPT, among all 1000 problem groups. In other words, Figure~\ref{fig:best_group} shows the performance of SPT and CASS for the single problem group with the greatest difference between a CASS configuration and SPT. In this single group, CASS achieves the maximum improvement of more than 30\% over SPT for this problem group on two of its configurations. We note that, since we tested 216 CASS configurations across 1000 different problem groups, there is a reasonable chance of observing such a large difference \emph{even if CASS performed identically to SPT in expectation.} We do not intend for this result to demonstrate statistical significance, but simply to illustrate the outcome of our experiments. 

Figure~\ref{fig:ap_mean} compares the mean of AP over all 1000 problem groups. In it, the blue bars, moving left to right, depict the CASS configurations that are \emph{(i)} the two best, \emph{(ii)} the median, and \emph{(iii)} the two worst in terms of average precision. Aroma's baseline SPT configuration is highlighted in orange. The best two CASS configurations show an average improvement of more than 1\% over SPT, while the others degraded performance relative to the baseline SPT configuration. 

These results illustrate that certain CASS configurations can outperform the SPT on average by a small margin, and can outperform the SPT on specific problem groups by a large margin. However, we also note that choosing a good CASS configuration for a domain is essential. We leave automating this configuration selection to future work.

\subsubsection{Analysis of Configurations}

\begin{figure}[ht]
\centering
\begin{subfigure}[b]{0.32\linewidth}
\includegraphics[width=\linewidth]{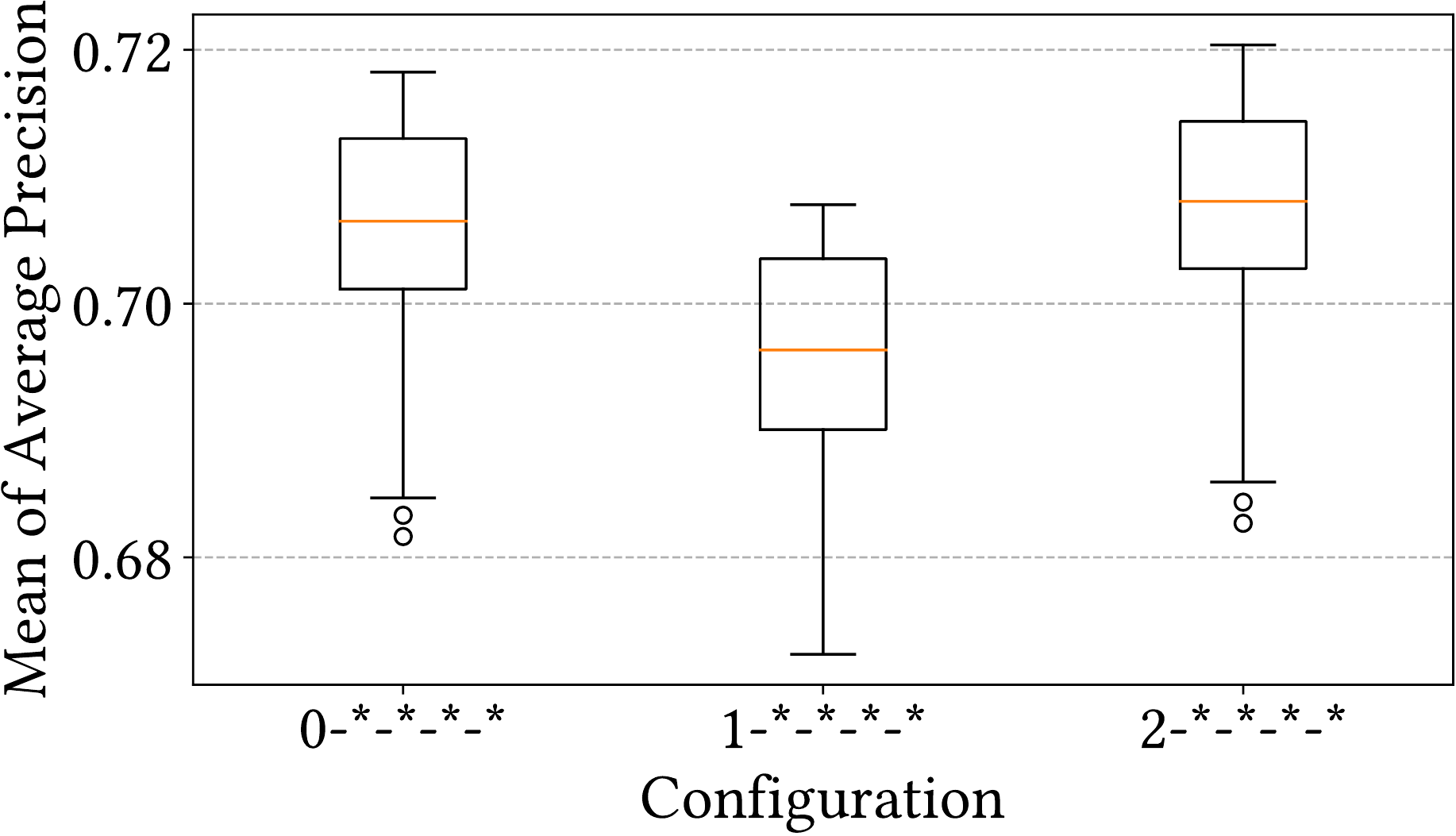}
\caption{Node Prefix Labels.}
\label{fig:c0}
\end{subfigure}
\hfill
\begin{subfigure}[b]{0.32\linewidth}
\includegraphics[width=\linewidth]{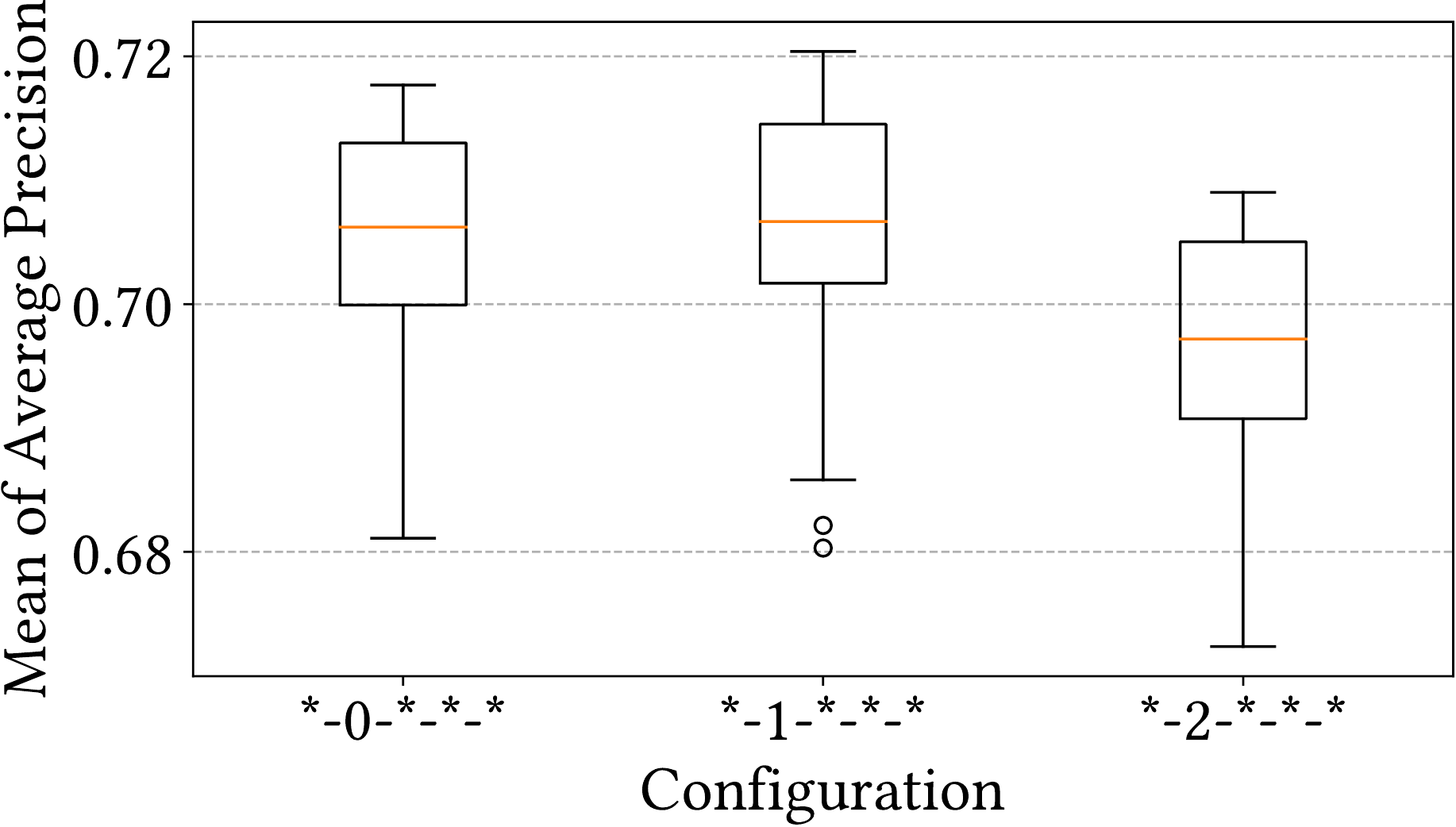}
\caption{Compound Statements.}
\label{fig:c1}
\end{subfigure}
\hfill
\begin{subfigure}[b]{0.32\linewidth}
\includegraphics[width=\linewidth]{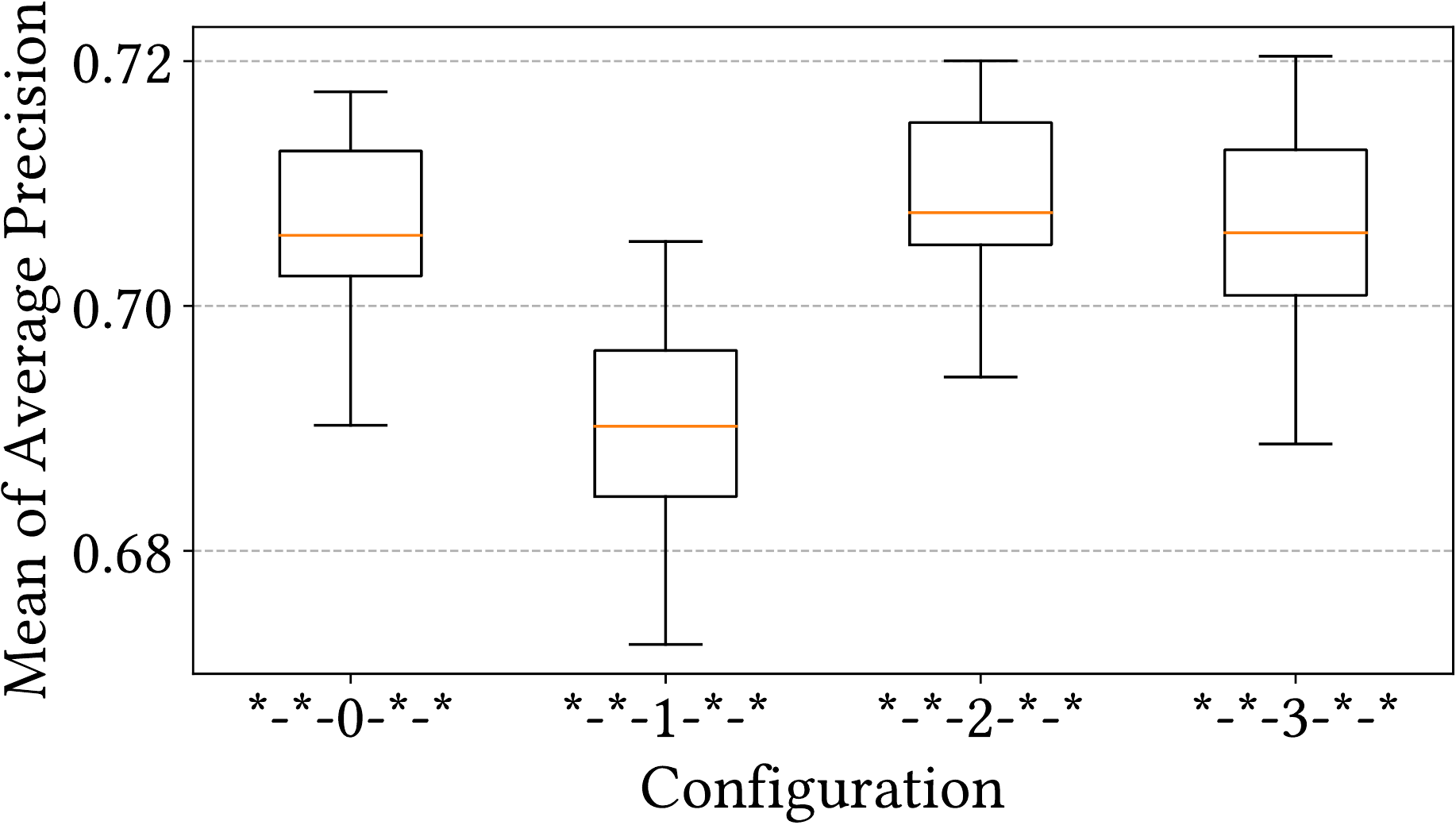}
\caption{Global Variables.}
\label{fig:c2}
\end{subfigure}
\par\medskip
\begin{subfigure}[b]{0.32\linewidth}
\includegraphics[width=\linewidth]{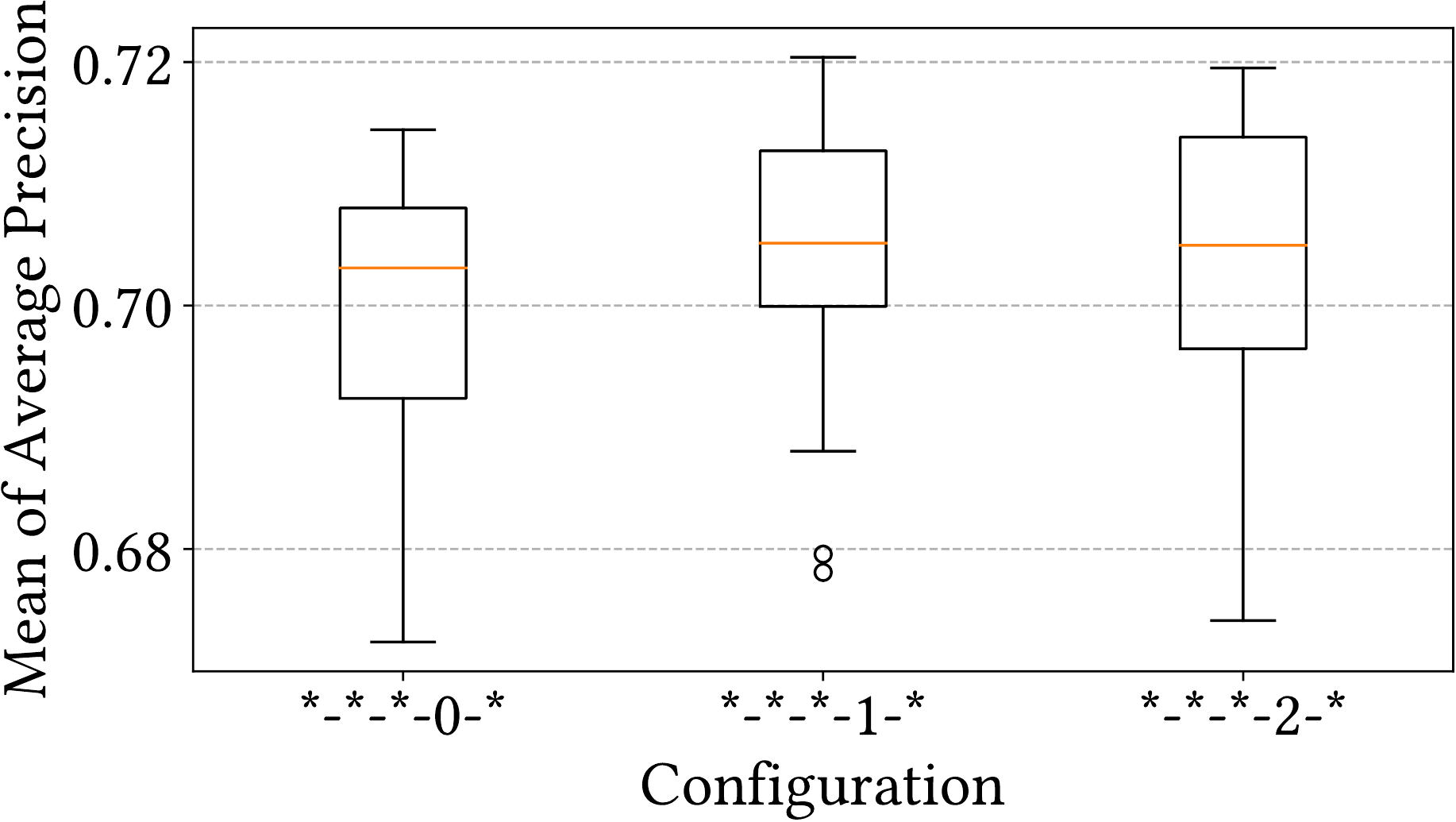}
\caption{Global Functions.}
\label{fig:c3}
\end{subfigure}
\quad
\begin{subfigure}[b]{0.32\linewidth}
\includegraphics[width=\linewidth]{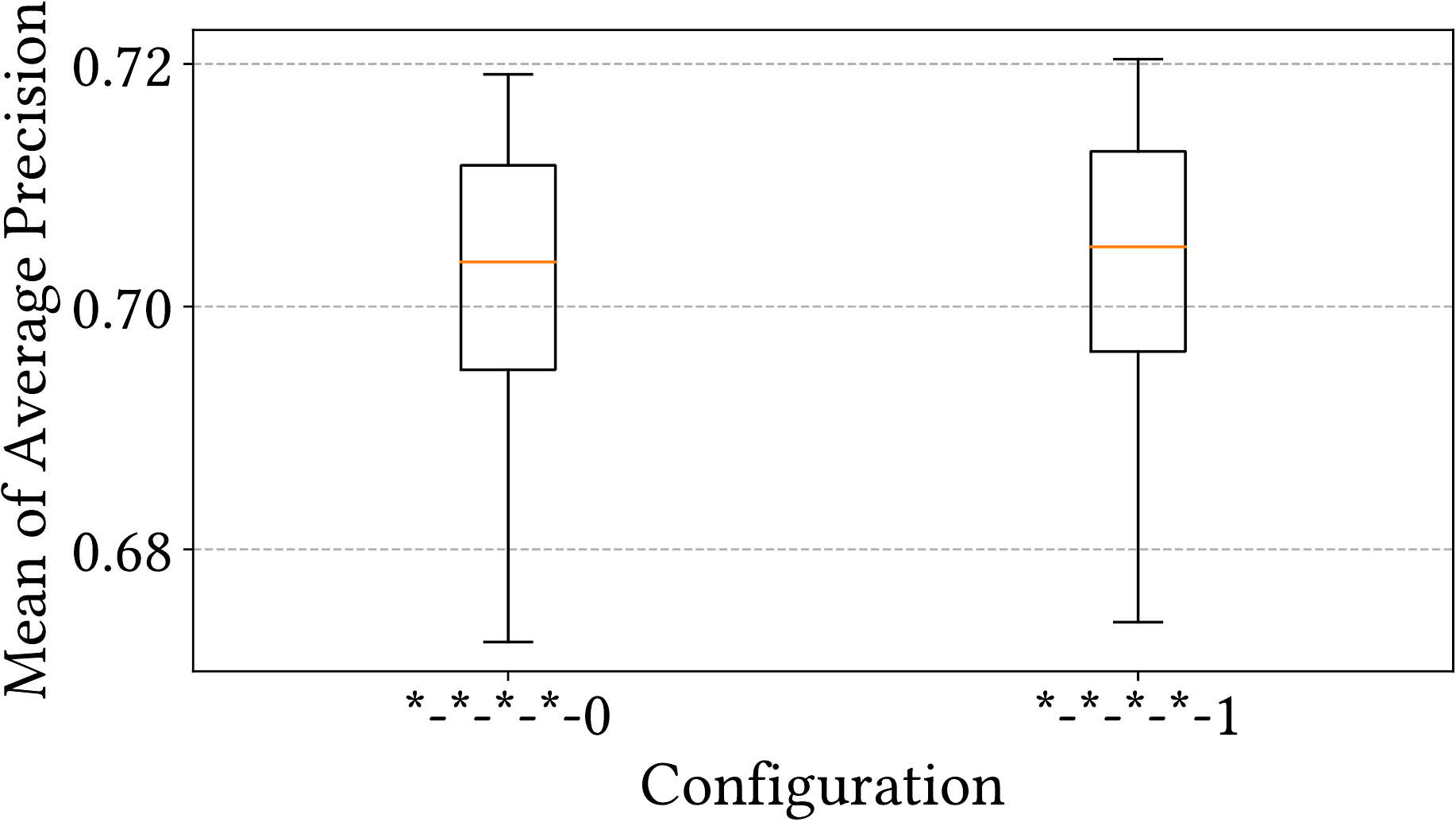}
\caption{function I/O cardinality.}
\label{fig:c4}
\end{subfigure}
\caption{The Distributions of Performance for Configurations with a Fixed Option Type.}
\label{fig:ap_fix_option}
\end{figure}

Figures~\ref{fig:c0}-\ref{fig:c4} serve to illustrate the performance variation for individual configurations. Figure~\ref{fig:c0} shows the effect of varying the options for the \textit{node prefix label} configuration. Applying the node prefix label for the parentheses operator (option 2) results in the best overall performance while annotating every internal node (option 1) results in a concrete syntax tree and the worst overall performance. This underscores the trade-offs in incorporating syntax-binding transformations in CASS. In Figure~\ref{fig:c1} we observe that removing all features relevant to \textit{compound statements} (option 1) leads to the best overall performance when compared with other options. This indicates that adding separate features for compound statements obscures the code's intended semantics when the constituent statements are also individually featurized. 

Figure~\ref{fig:c2} shows that removing all features relevant to \textit{global variables} (option 1) degrades performance. We also observe that eliminating the global variable identifiers and assigning a label to signal their presence (option 2) performs best overall, possibly because global variables appearing in similar contexts may not use the same variable identifiers. Further, option 2 performs better than the case where global variables are indistinguishable from local variables (option 3).  Figure~\ref{fig:c3} indicates that removing features relevant to identifiers of \textit{global functions}, but flagging their presence with a special label as done in option 2, generally gives the best performance. This result is consistent with the intuitions for eliminating features of function identifiers in CASS as discussed in Section~\ref{subsec:intuit}. Figure~\ref{fig:c4} shows that capturing the input and output cardinality improves the average performance. This aligns with our assumption that function I/O cardinality may abstract the semantics of certain groups of functions.

\paragraph{A Subtle Observation.}
A more nuanced and subtle observation is that our results seem to indicate that for each CASS configuration the optimal granularity of abstraction detail is different. For \textit{compound statements}, the best option seems to correspond to the coarsest level of abstraction detail, while for \textit{node prefix label}, \textit{global variables}, and \textit{global functions} the best option seems to corresponds to one of the intermediate levels of abstraction detail. Additionally, for \textit{function I/O cardinality}, the best option has a finer level of detail. For our future work, we aim to perform a deeper analysis on this and hopefully learn such configurations, to reduce (or eliminate) the overhead necessary of trying to manually discover such configurations. 

\section{Broader Impact} \label{sect:impact}
 
To discuss the broader impact of our project, we will categorize impacts by their degree of influence. For example, by the phrase ``first-degree negative impact'' we will refer to a scenario where a given research idea can be directly used for harm (e.g., DeepFake~\citep{floridi:2018:pt}, DeepNude\footnote{We have intentionally not included a citation to this work. We do not want to be seen, in any way, as endorsing or promoting it. We believe such an act would be ethically irresponsible.}, so on).  Similarly, by "second-degree negative impact" we will refer to a scenario where a research idea may have a direct negative or positive impact based on how it is used (e.g., facial recognition for security vs. oppressing minorities, GPT~\citep{radford:2019:openai} to create an empathetic chatbot vs. malicious fake news, etc). We call a research idea to have a "third-degree negative impact" if the idea by itself represents an abstract concept (e.g., a similarly metric) and cannot harm by its own, but can be used to build a second application which can then have a negative impact based on its use.

We envision the following positive broader impacts of the research idea presented in this paper. As briefly mentioned in the introduction, an end-to-end code similarity system can be incorporated in programming tools (e.g., Visual Studio, Eclipse, etc.) to improve the productivity of a programmer by offering him/her a similar but "known to be more efficient" code snippet. It can be used in coding education by displaying better (e.g., concise, faster, space-efficient, etc.) code for a given code snippet,  in assisting program debugging by identifying potential missing parts, for plagiarism detection, for automated bug-detection and fixing, in automatic code transformations (e.g., replacing a Python function with an equivalent C function) and so on. If used wisely with proper control and governance, we believe it can create many positive impacts.

We can envision the following third-degree negative impacts. If a tool that uses code similarity becomes mature enough to automatically generate correct compilable codes, it can be potentially used to automatically replace code from one language to another or to replace a slow code with a fast one. A malicious person can leverage the code similarity tool to crawl the web and steal codes on the web, find common patterns and security flaws in the code available on the web, and then find ways to hack at a massive scale. Codes generated from the same code generators are likely to be more vulnerable to such attacks. If systems allow automatic code patching/fixing based on code-similarity without proper testing, it might create security flaws if hacked. If programmers get used to getting help from a programming tool, that might negatively reduce the learning ability of programmers unless the tool also offers explainability. Explainability would be required to understand what the tool is learning about the code similarity and to educate the programmers about it.

To summarize, code similarity is an abstract concept that is likely to have numerous positive applications. However, if used in other tools, it might also play a role in creating a third-order negative impact. It may be used to develop tools and applications which, if mature enough, may cause unacceptable or dangerous situations. To mitigate the negative impacts, we would need to ensure proper policy and security measures are in place to prevent negative usage. In particular, such secure systems may require a human-in-the-loop so that any such tool is used to enhance the capability and productivity of programmers.

\end{document}